\documentclass[Afour,sageh,times]{sagej}

\usepackage{algorithm}
\usepackage{algpseudocode}
\usepackage{graphicx}
\usepackage{textcomp}
\usepackage{xcolor}
\usepackage{enumitem}
\usepackage{romannum}
\usepackage{gensymb}
\usepackage{multirow}
\usepackage{tabularx}
\usepackage{url}
\usepackage{fancyhdr}
\usepackage[bookmarksopen,bookmarksnumbered]{hyperref}

\newcommand\BibTeX{{\rmfamily B\kern-.05em \textsc{i\kern-.025em b}\kern-.08em
T\kern-.1667em\lower.7ex\hbox{E}\kern-.125emX}}

\begin{document}

\runninghead{Khandelwal N. \textit{et al.}}

\title{Distributed Inverse Dynamics Control for Quadruped Robots using Geometric Optimization\\}

\author{
Nimesh Khandelwal\affilnum{1,2}, Amritanshu Manu\affilnum{1},\\ Shakti S. Gupta\affilnum{1}, Mangal Kothari\affilnum{1}, Prashanth Krishnamurthy\affilnum{2}, Farshad Khorrami\affilnum{2}
}

\affiliation{\affilnum{1}Deptartment of Mechanical Engineering, IIT Kanpur, Kanpur, UP, India\\
\affilnum{2}Department of Electrical and Computer Engineering, New York University, Brooklyn, NY,  USA
}

\corrauth{Nimesh Khandelwal,
Mobile Robotics Laboratory,
Department of Mechanical Engineering,
IIT Kanpur, Kanpur, UP,
208016, India.}

\email{nimesh20@iitk.ac.in}

\begin{abstract}
   This paper presents a distributed inverse dynamics controller (DIDC) for quadruped robots that addresses the limitations of existing reactive controllers: simplified dynamical models, the inability to handle exact friction cone constraints, and the high computational requirements of whole-body controllers. Current methods either ignore friction constraints entirely or use linear approximations, leading to potential slip and instability, while comprehensive whole-body controllers demand significant computational resources. Our approach uses full rigid-body dynamics and enforces exact friction cone constraints through a novel geometric optimization-based solver. DIDC combines the required generalized forces corresponding to the actuated and unactuated spaces by projecting them onto the actuated space while satisfying the physical constraints and maintaining orthogonality between the base and joint tracking objectives. Experimental validation shows that our approach reduces foot slippage, improves orientation tracking, and converges at least two times faster than existing reactive controllers with generic QP-based implementations. The controller enables stable omnidirectional trotting at various speeds and consumes less power than comparable methods while running efficiently on embedded processors.
\end{abstract}


\maketitle

\section{Introduction}
\label{sec:intro}
Quadruped robots are being deployed in the real world for autonomous inspection, mapping, and surveillance operations in different environments. This has been realized by developing versatile control algorithms suitable for fast execution on limited onboard computational resources. The locomotion control algorithms can be broadly classified into model-based (e.g., {\cite{cheetah3, imc, vmc-slope, convex-mpc, nmpc-farbod}}), and learning-based algorithms (e.g., \cite{learning1, learning2, learning3}). A current trend in state-of-the-art model-based control implementations is to use a combination of predictive and reactive controllers (e.g., \cite{nmpc-neunert, wbic-mit, sleiman}). In this setup, the predictive controller uses a simplified model of the robot and its interactions with the environment to provide the reference state trajectory and control inputs for a time horizon. The reactive controller uses a more accurate dynamics model, enforces state and control constraints, and prioritizes different tracking tasks. Thus, the reactive layer generates physically consistent joint torque commands at each time instant and allows the robot to respond quickly to unforeseen disturbances. 

Reactive controllers for quadruped robots have been explored extensively in the literature in the last two decades. The initial approach was motivated by the inverse dynamics control (IDC) of manipulators (\cite{robot-book}). However, in the case of quadruped robots, the under-actuated dynamics, dynamically switching contact states, and unknown contact forces proved to be a challenge. This was resolved by eliminating the contact force from the equation of motion of the system by using null-space projection and orthogonal decomposition of the contact Jacobian (e.g. \cite{idc_qr, idc_qr2, idc_qr3}). Although the classic inverse dynamics control provides an analytical form for the joint torques, it does not ensure that the contact forces generated by these torques satisfy the unilaterality and friction cone constraints. This limitation was overcome by the introduction of the quadratic program (QP) based methods for finding the optimal distribution of the contact forces (\cite{cheetah3}, \cite{vmc2013}). Using the single rigid body dynamics (SRBD), a linear approximation of the friction cone (as a friction pyramid), and mature open-source QP solvers (like qpOASES (\cite{qpoases}), OSQP (\cite{osqp}), etc.), these methods are suitable for onboard implementations.

Subsequently, more comprehensive QP formulations have been developed that use the full rigid body dynamics (RBD), simultaneously optimize over joint torques, joint accelerations, and contact forces, and achieve prioritized constraint satisfaction. These are collectively known as Whole Body Controllers (WBC) (e.g. \cite{wbc, wbc-fahmi, wbc-simple}). The prioritized constraint satisfaction in the WBC can be handled either by formulating a single QP with weighted objectives or by formulating multiple QPs that are solved hierarchically. However solving the WBC problem at a fast enough rate does require a high-end onboard processor, for instance, the ANYmal robot (\cite{anymal}) runs WBC at 400 Hz using an Intel Core i7-8850H processor (\cite{sleiman}). This is primarily due to the increased size of the problem due to multiple constraints. Additionally, to keep the optimization problem as a QP, the WBC formulations need to use the friction pyramid approximation. In (\cite{qcqp-aghili}), the proposed controller uses the full-RBD model and the exact friction cone to form a QCQP, which is solved using an interior-point algorithm. However, an onboard implementation of the controller for a quadruped robot is not provided. Thus, it is of value to develop a reactive controller that uses accurate system dynamics and exact friction constraints while running comfortably on an embedded processor (e.g. ARM64 processors).

In this work, we propose a reactive controller, called distributed inverse dynamics controller (DIDC), that uses the full-RBD model of a quadruped robot and enforces the exact friction cone constraint. We also propose a geometric-optimization-based custom solver that keeps the controller computationally lightweight. The optimization problem involved in our controller is inspired by (\cite{cheetah3}), and our custom solver to handle the cone constraint leverages the projection operator proposed in (\cite{friction_projection}) in the context of friction cones. We develop an approach to project the generalized forces required for unactuated degrees of freedom (DOFs) into the space of actuated DOFs while enforcing the cone constraints. The DIDC also maintains the orthogonality of the control actions for the unactuated and the actuated space so that the tracking of base motion is unaffected by joint tracking. This work also includes a control architecture using the DIDC along with a heuristic planner for the base-motion and footholds, and an extended Kalman filter (EKF) based state estimator. The proposed control architecture is validated for omnidirectional trotting at various speeds, in simulation and on hardware, using the Go2 quadruped robot from Unitree Robotics.

In the following, we give an overview of the RBD and the IDC in Section \ref{sec:idc}. We describe the proposed reactive controller and the geometric-optimization-based solver in Section \ref{sec:didc} and Section \ref{sec:optim}, respectively. The motion planner and the state estimator modules of the control architecture are discussed in Section \ref{sec:p&e} followed by the experimental results in Section \ref{sec:experiments}.

\section{Reactive Control of Quadruped Robots}
\label{sec:idc}
\subsection{Model formulation}
The DOFs of the robot are represented by the generalized coordinates vector $\mathbf{q}\in\mathbb{R}^{18}$. Since the system is underactuated, the DOFs are partitioned into the 6-DOF unactuated space for the robot base, and the 12-DOF actuated space for the leg joints, as $\mathbf{q}=\begin{bmatrix}
    \mathbf{q}_b^T&\mathbf{q}_j^T
\end{bmatrix}^T$. Here, $\mathbf{q}_b=\begin{bmatrix}
    \mathbf{r}_b^T&\mathbf{\Phi}_b^T
\end{bmatrix}^T\in \mathbb{R}^6$, where $\mathbf{r}_b\in\mathbb{R}^{3}$ is the position vector of the robot center-of-mass (COM) and $\mathbf{\Phi}_b\in \mathbb{R}^3$ is the \textit{XYZ}-Euler angle vector representing the orientation of the robot base. $\mathbf{q}_j\in\mathbb{R}^{12}$ is the vector of joint angles for all four legs, ordered as front-left (FL), front-right (FR), rear-left (RL), and rear-right (RR), with each leg numbered 0, 1, 2, and 3 respectively. 

The general equation of motion (EOM) of a quadruped robot, modeled as a rigid-body system with contact constraints, is expressed as
\begin{equation}
    \mathbf{M}(\mathbf{q})\ddot{\mathbf{q}} + \boldsymbol{\eta}(\mathbf{q}, \dot{\mathbf{q}}) = \mathbf{S}^T\boldsymbol{\tau} + \mathbf{J}_c^T\mathbf{F}_c,
    \label{eq:eom}
\end{equation}
where $\mathbf{M}(\mathbf{q})\in\mathbb{R}^{18\times 18}$ is the joint space inertia matrix of the system, and $\boldsymbol{\eta}(\mathbf{q}, \dot{\mathbf{q}})\in\mathbb{R}^{18}$ represents the nonlinear terms corresponding to Coriolis, centrifugal and gravitational effects. $\mathbf{S} = \begin{bmatrix}
    \mathbf{0}_{12\times6}&\mathbf{I}_{12\times 12}
\end{bmatrix}$ is the selection matrix representing the underactuation since only the joint DOFs are actuated. $\boldsymbol{\tau}\in\mathbb{R}^{12}$ is the joint torque command, $n_c$ is the number of feet in contact, $\mathbf{J}_c\in\mathbb{R}^{3n_c\times 18}$ is the contact Jacobian matrix, and $\mathbf{F}_c\in\mathbb{R}^{3n_c}$ are the contact forces on the feet of the robot. The contact constraint enforces the no-slip condition at the point of contact between the foot and the environment. It is expressed mathematically as
\begin{equation}
\begin{aligned}
    \mathbf{J}_{c,i}\dot{\mathbf{q}} &= \mathbf{0},\\
    \mathbf{J}_{c,i}\ddot{\mathbf{q}}+\dot{\mathbf{J}}_{c,i}\dot{\mathbf{q}}&=\mathbf{0},
\end{aligned}
\end{equation}
where $\mathbf{J}_{c,i}$ and $\dot{\mathbf{J}}_{c,i}$ are the sub-matrix of the contact jacobian, $\mathbf{J}_c$, and its time derivative, $\dot{\mathbf{J}}_{c,i}$, corresponding to feet $i \in \mathcal{C}$, where $\mathcal{C}$ is the set of feet in contact with the ground. For physical correctness, the feet in contact should only push into the environment and the contact forces must satisfy the friction constraints. These are together referred to as contact force constraints. Their mathematical expression is discussed in Equations (\ref{eq:unilateral_force}) and (\ref{eq:friction_cone}) in Section \ref{sec:didc}.

The physical quantities used in the planner, estimator, and controller (PEC) computations are expressed either in the global (inertial) frame $\mathcal{G}$ or the local frame  $\mathcal{B}$ that is attached to the COM and has the same orientation as the robot base. The axes of these frames are denoted with a superscript describing the frame, e.g. $X^{\mathcal{G}}$, whereas the notation $^{\mathcal{B}}_{}\mathbf{r}_b$ implies that $\mathbf{r}_b$ is expressed in the frame $\mathcal{B}$.

\subsection{Null-space projection inverse dynamics control (NSPIDC)}
As described in Section \ref{sec:intro}, projecting the EOM from (\ref{eq:eom}) into the null space of the contact constraints using a dynamically consistent null-space projection matrix was the initial approach toward designing reactive controllers for quadruped robots. This renders the resulting dynamics invariant to contact forces. The joint torque command $\boldsymbol{\tau}$ is calculated as
\begin{equation}
    \boldsymbol{\tau} = (\mathbf{N}_c^T\mathbf{S}^T)^{\dagger}\mathbf{N}_c^T(\mathbf{M}\ddot{\mathbf{q}}+\boldsymbol{\eta})
    \label{eq:nsp}
\end{equation}
where $\mathbf{N}_c = \mathbf{I}_{18\times 18} - \mathbf{M}^{-1}\mathbf{J}_c^T(\mathbf{J}_c\mathbf{M}^{-1}\mathbf{J}_c^T)^{-1}\mathbf{J}_c$ is the dynamically consistent null-space projection matrix of contact constraints. Since the resulting inverse dynamics (\ref{eq:nsp}) is contact invariant, the joint torques $\boldsymbol{\tau}$ do not compensate for the contact forces on the feet. Additionally, the contact forces generated at the feet using this controller have no guarantee of satisfying the friction constraints.

\subsection{QP-based balance control (BC)}
A prevalent class of reactive controllers, as described in Section \ref{sec:intro}, use simplified dynamics models and formulate a constrained QP to solve for desired contact forces acting at the stance legs, and use only the leg dynamics to compute the desired joint torques for the swing legs. Although these controllers compensate for the total mass of the robot while satisfying the linearized friction constraints, they fail to account for the effect of leg dynamics on the base. Also, the optimal torques required for base motion are modified by the direct addition of torques required for swing leg motion, and base motion tracking.

The shortcomings of NSPIDC, BC, and the related family of reactive controllers are overcome by the control formulation described in this work.

\section{Distributed Inverse Dynamics Control}
\label{sec:didc}
The DIDC uses the dynamics model in (\ref{eq:eom}) to calculate joint torque command $\boldsymbol{\tau}$ while satisfying contact constraints and the friction cone constraint. The key ideas of the DIDC are summarized below:
\begin{enumerate}
    \item considering the system as fully-actuated and computing the required generalized forces $\boldsymbol{\tau}_f = \begin{bmatrix}
    \boldsymbol{\tau}_b^T&\boldsymbol{\tau}_j^T
\end{bmatrix}^T \in  \mathbb{R}^{18}$,
    \item mapping the generalized forces for the unactuated space $\boldsymbol{\tau}_b \in \mathbb{R}^{6}$ to the control input in the actuated space (expressed here as $\boldsymbol{\tau}_1 \in \mathbb{R}^{12}$), while enforcing the contact and friction constraints (\ref{eq:tau_1}),
    \item and adding the generalized forces for the actuated space $\boldsymbol{\tau}_j \in \mathbb{R}^{12}$ to $\boldsymbol{\tau}_1$, while maintaining the orthogonality of the control actions, to get $\boldsymbol{\tau}$ (\ref{eq:final_tau}). 
\end{enumerate}
In the following, we describe in detail each of these steps in the computation of $\boldsymbol{\tau}$. We begin by expressing the general EOM in (\ref{eq:eom}) with the introduction of $\boldsymbol{\tau_f}$ as 
\begin{equation}
    \label{eq:tau_full}
    \mathbf{M}(\mathbf{q})\ddot{\mathbf{q}} + \boldsymbol{\eta}(\mathbf{q}, \dot{\mathbf{q}}) = \boldsymbol{\tau_f},
\end{equation}
assuming all the DOFs are actuated and no external forces are acting at the feet. Given the commanded generalized acceleration $\ddot{\mathbf{q}}_{cmd}$, we can calculate $\boldsymbol{\tau}_f$ as
\begin{align}
    \label{eq:tau_est}
    \boldsymbol{\tau}_f = \hat{\mathbf{M}}(\mathbf{q})\ddot{\mathbf{q}}_{cmd}+\hat{\boldsymbol{\eta}}(\mathbf{q}, \dot{\mathbf{q}}),
\end{align}
where the $\hat{(.)}$ denotes the available estimates of the corresponding quantities of the system. Substituting the value of $\boldsymbol{\tau}_f$ in (\ref{eq:tau_full}) using (\ref{eq:tau_est}), we express $\ddot{\mathbf{q}}$ as
\begin{equation}
    \ddot{\mathbf{q}} = \mathbf{M}^{-1}(\hat{\mathbf{M}}\ddot{\mathbf{q}}_{cmd} + \Tilde{\boldsymbol{\eta}}),
    \label{eq:qdd}
\end{equation}
where $\Tilde{\boldsymbol{\eta}} = \hat{\boldsymbol{\eta}} - \boldsymbol{\eta}$, and we have dropped the explicit dependence on $\mathbf{q}$ and $\ddot{\mathbf{q}}$ for notational simplicity. With further manipulation of (\ref{eq:qdd}), we arrive at a partially-linearized dynamics of the form 
\begin{equation}
\begin{aligned}
    \label{eq:lin_dyn}
    \ddot{\mathbf{q}} &= \ddot{\mathbf{q}}_{cmd} + \mathbf{M}^{-1}(\hat{\mathbf{M}}\ddot{\mathbf{q}}_{cmd} + \Tilde{\boldsymbol{\eta}}) - \mathbf{M}^{-1}\mathbf{M}\ddot{\mathbf{q}}_{cmd}, \\
    \ddot{\mathbf{q}} &= \ddot{\mathbf{q}}_{cmd}+
    \boldsymbol{\epsilon}(\mathbf{q}, \dot{\mathbf{q}}, \ddot{\mathbf{q}}_{cmd}),
\end{aligned}
\end{equation}
where the quantity $\boldsymbol{\epsilon}(\mathbf{q}, \dot{\mathbf{q}}, \ddot{\mathbf{q}}_{cmd})$ is the modeling uncertainty and is defined as
\begin{align*}
    \boldsymbol{\epsilon}(\mathbf{q}, \dot{\mathbf{q}}, \ddot{\mathbf{q}}_{cmd}) = \mathbf{M}^{-1}(\Tilde{\mathbf{M}}\ddot{\mathbf{q}}_{cmd} + \Tilde{\boldsymbol{\eta}}),
\end{align*}
with $\Tilde{\mathbf{M}} = \hat{\mathbf{M}} - \mathbf{M}$. If we assume no modeling uncertainty, the dynamics in (\ref{eq:lin_dyn}) reduces to a linear ODE
\begin{equation}
    \ddot{\mathbf{q}} = \ddot{\mathbf{q}}_{cmd}.
\end{equation}
This is the decoupled-linearized dynamics of the system since each DOF can be controlled individually by their corresponding commanded acceleration, $\ddot{\mathbf{q}}_{cmd}$, which is calculated as
\begin{equation} \ddot{\mathbf{q}}_{cmd}=\ddot{\mathbf{q}}_{des}+\mathbf{K}_p(\mathbf{q}_{des}\ominus \mathbf{q})+\mathbf{K}_d(\dot{\mathbf{q}}_{des}-\dot{\mathbf{q}}),
\end{equation}
where $\mathbf{K}_p$ and $\mathbf{K}_d$ are the stiffness and damping coefficient matrices, chosen to be diagonal here, and $(.)_{des}$ denotes the desired quantities provided by the motion planner. The symbol $\ominus$ denotes the space-preserving subtraction operator. It differs from the algebraic subtraction operator $(-)$ for the orientation coordinates since they belong to a non-Euclidean space. The $\ominus$ operator is used here to calculate the error in the orientation space and represent it in $\mathbb{R}^3$.

To control the 18-DOF system, $\boldsymbol{\tau}_f$ needs to be distributed over the actuated space to calculate the joint torque command. Since the robot interacts with the environment through its feet, it can produce the desired base wrench , $\boldsymbol{\tau}_b$, only by manipulating the contact forces , $\mathbf{F}_c$, at its feet. These are related by the sub-matrix $\mathbf{J}_{ab}$ of the contact Jacobian $\mathbf{J}_c = \begin{bmatrix}
    \mathbf{J}_{ab}&\mathbf{J}_{aa}
\end{bmatrix}$ as
\begin{equation}
    \mathbf{J}_{ab}^T\mathbf{F}_c=\boldsymbol{\tau}_b.
    \label{eq:base_tau}
\end{equation}
The matrix, $\mathbf{J}_{ab}^T\in\mathbb{R}^{6\times3n_c}$, has the following structure
\begin{equation}
    \mathbf{J}_{ab}^T=\begin{bmatrix}
    \mathbf{I}_{3\times 3} & ... & \mathbf{I}_{3\times 3}\\
    (\mathbf{r}_{p_i}-\mathbf{r}_b)^{\times} & ... & (\mathbf{r}_{p_{n_c}}-\mathbf{r}_b)^{\times}
    \end{bmatrix},
\end{equation}
where $\mathbf{r}_{p_i}\forall\; i\in\{1, 2,\ldots, n_c\}$ is the position of the feet in contact, $\mathbf{r}_b$ is the position of the COM. The superscript $(\times)$ in $\mathbf{r}^\times \in \mathfrak{so}(3) \text{ such that } (\mathbf{r}^{\times})^T=-\mathbf{r}^{\times}$ represents the skew-symmetric matrix form of the vector $\mathbf{r}\in\mathbb{R}^3$. Note that (\ref{eq:base_tau}) is similar to existing QP-based controller formulations in (\cite{cheetah3}) and (\cite{vmc2013}). The main difference here is that the term $\boldsymbol{\tau}_b$ accounts for the complete dynamics of the system, including the interactions between different links of the quadruped robot, instead of the simplified dynamics (like SRBD or centroidal dynamics). Specifically, the nonlinear effects $\boldsymbol{\eta}(\mathbf{q}, \dot{\mathbf{q}})$ have a non-negligible contribution on the desired base wrench. 

To solve Equation (\ref{eq:base_tau}), we formulate it as a quadratic optimization problem with regularization
\begin{equation}
\begin{aligned}
\mathbf{F}_c^* = \arg \min_{\mathbf{F}_c} \quad & (\mathbf{J}_{ab}^T\mathbf{F}_c-\boldsymbol{\tau}_b)^T\mathbf{S}_1(\mathbf{J}_{ab}^T\mathbf{F}_c-\boldsymbol{\tau}_b) \\
                                               & + \mathbf{F}_c^T\mathbf{W}\mathbf{F}_c \\
                                               & + (\mathbf{F}_c - \mathbf{F}^*_{c,prev})^T\mathbf{V}(\mathbf{F}_c - \mathbf{F}^*_{c,prev}) \\
\text{subject to} \quad                        & \mathbf{g}(\mathbf{F}_c) \leq \mathbf{0},
\end{aligned}
\label{eq:optimization}
\end{equation}
where $\mathbf{g}(\mathbf{F}_c)$ represents the contact constraints stacked as a single vector. $\mathbf{F}_{c,prev}^*$ is the solution from the previous time step. The second and third terms in the cost function penalize large contact forces and large contact impulses, respectively. The matrix, $\mathbf{S}_1$, is the relative weighting matrix of the resulting base wrench. Its values are chosen to prioritize base orientation stabilization over base translation motion. The matrices, $\mathbf{W}$, and, $\mathbf{V}$, define the regularization of contact forces and impulses. The solution of this quadratic problem is subject to the unilaterality and maximum normal force constraint 
\begin{equation}
    F^{z}_{c,\text{max}}\ge F_c^z\geq F^{z}_{c,\text{min}} \geq 0,
    \label{eq:unilateral_force}
\end{equation}
along with friction cone constraint 
\begin{equation}
    \sqrt{(F_c^x)^2+(F_c^y)^2}\le\mu |F_c^z|,
    \label{eq:friction_cone}
\end{equation}
represented here as a single constraint vector $\mathbf{g}(\mathbf{F}_c)$. $\mu$ is the coefficient of friction. The optimization algorithm used to solve (\ref{eq:optimization}) is explained in Section \ref{sec:optim}.

The solution of this quadratic optimization is the required forces on the feet in contact, $\mathbf{F}_c^*$. We can get the required joint torque command $\boldsymbol{\tau}_1$ for tracking the base motion from $\mathbf{F}_c^*$ using the static equilibrium relation for the stance legs
\begin{equation}
\boldsymbol{\tau}_1 = -\mathbf{J}_{aa}^T\mathbf{F}_c^*.
\label{eq:tau_1}
\end{equation}
To ensure tracking for all the leg joints, $\boldsymbol{\tau}_j$ is the required additional joint torque. However, a direct addition of $\boldsymbol{\tau}_1$ and $\boldsymbol{\tau}_j$ will modify the resulting wrench on the base as the joint torque commands for the stance legs will change. Here, we prioritize tracking of the base motion over joints by projecting $\boldsymbol{\tau}_j$ into the null space of the Jacobian $\mathbf{J}_{a,b}$ mapping the base wrench $\boldsymbol{\tau}_b$ to joint torques. 
We use a composition of Jacobians to define $\mathbf{J}_{a,b} = -\mathbf{J}_{aa}^T(\mathbf{J}_{ab}^T)^{\dagger}$. Thus, for a given total joint torque vector, $\boldsymbol{\tau}$, the effective wrench on the base can be written as
\begin{equation}
    \hat{\boldsymbol{\tau}}_b=(\mathbf{J}_{a,b})^\dagger \boldsymbol{\tau}.
    \label{eq:est_base_wrench}
\end{equation}

To understand this mapping, it is helpful to consider that if we neglect the contact constraints, Equation (\ref{eq:base_tau}) can be solved for $\mathbf{F}_c^*$ by taking the pseudo-inverse of $\mathbf{J}_{ab}^T$ and multiplying it with $\boldsymbol{\tau}_b$. The resulting contact forces multiplied by $-\mathbf{J}_{aa}^T$ would give us the base tracking torques. The solution we get from the optimization is an element of the subset of this solution space, defined by the contact constraints. 

Since $\mathbf{J}_{a,b}$ defines the mapping from the body wrench to joint torques, the additional torques required for joint tracking should lie in the null space of $\mathbf{J}_{a,b}$ so that the base motion tracking is not affected. The corresponding null-space matrix is defined as
\begin{equation}
\begin{aligned}
    \mathbf{N}_{a,b} &=\mathbf{I}_{12\times 12} - \mathbf{J}_{a,b}(\mathbf{J}_{a,b})^\dagger\\
    &= \mathbf{I}_{12\times 12} - \mathbf{J}_{a,b}(\mathbf{J}_{a,b}^T\mathbf{J}_{a,b})^{-1}\mathbf{J}_{a,b}.
\end{aligned}
\end{equation}
 The final joint torque command is calculated as follows
 \begin{equation}
     \boldsymbol{\tau} = -\mathbf{J}_{aa}^T\mathbf{F}^*_c + \mathbf{N}_{a,b}\boldsymbol{\tau}_j.
     \label{eq:final_tau}
 \end{equation}

 The null space projection ensures that the base and joint motion tracking components of $\boldsymbol{\tau}$ are always orthogonal. This can be verified by substituting the value of $\boldsymbol{\tau}$ from (\ref{eq:final_tau}) in (\ref{eq:est_base_wrench}) to calculate the effective base wrench, neglecting the contact constraints, as shown
 \begin{align*}
     \hat{\boldsymbol{\tau}}_b&=(\mathbf{J}_{a,b})^\dagger \boldsymbol{\tau}\\
     &=(\mathbf{J}_{a,b})^\dagger(-\mathbf{J}_{aa}^T\mathbf{F}^*_c + \mathbf{N}_{a,b}\boldsymbol{\tau}_j)\\
     &=(\mathbf{J}_{a,b})^\dagger\mathbf{J}_{a,b}\boldsymbol{\tau}_b + (\mathbf{J}_{a,b})^\dagger\mathbf{N}_{a,b}\boldsymbol{\tau}_j,
 \end{align*}
 where the first term is the projection of $\boldsymbol{\tau}_b$ into the column space of $\mathbf{J}_{a,b}$. The second term simplifies to zero using the definition of the generalized inverse (\cite{generalized_inverse}) that for any matrix $\mathbf{A}$, a valid inverse $\mathbf{G}$ satisfies $\mathbf{A}\mathbf{G}\mathbf{A}=\mathbf{A}\Leftrightarrow\mathbf{G}\mathbf{A}\mathbf{G}=\mathbf{G}$.
 \begin{align*}
     (\mathbf{J}_{a,b})^\dagger\mathbf{N}_{a,b}&=(\mathbf{J}_{a,b})^\dagger(\mathbf{I}_{12\times 12} - \mathbf{J}_{a,b}(\mathbf{J}_{a,b}^T\mathbf{J}_{a,b})^{-1}\mathbf{J}_{a,b}^T)\\
     &= (\mathbf{J}_{a,b})^\dagger - (\mathbf{J}_{a,b})^\dagger\mathbf{J}_{a,b}(\mathbf{J}_{a,b}^T\mathbf{J}_{a,b})^{-1}\mathbf{J}_{a,b}^T\\
     &= (\mathbf{J}_{a,b})^\dagger - (\mathbf{J}_{a,b})^\dagger\mathbf{J}_{a,b}(\mathbf{J}_{a,b})^\dagger\\
     &=(\mathbf{J}_{a,b})^\dagger-(\mathbf{J}_{a,b})^\dagger\\
     &=\mathbf{0}.
 \end{align*}
  In our case, $(\mathbf{J}_{a,b})^\dagger\mathbf{J}_{a,b}(\mathbf{J}_{a,b})^\dagger=(\mathbf{J}_{a,b})^\dagger$. 
 This shows that both the objectives of base and joint tracking are satisfied independently and simultaneously. This is similar to the hierarchical control formulation (\cite{hierarchical}) if we consider the base control as the higher priority task and joint tracking as a lower priority task.
 
 We must take note that the above holds only until the system is fully- (or over-) constrained, which in the case of quadruped robots implies that at least two feet are in contact with the ground. This ensures that the matrix $\mathbf{J}_{ab}^T$ is always full row rank. If more than two feet are in contact with the ground ($n_c>2; \mathbf{J}_{ab}^T\in\mathbb{R}^{6\times (6+m)}; m\in\{3, 6\}$), then there are infinite solutions to (\ref{eq:base_tau}) and the final solution of the QP is the one that minimizes the cost function and satisfies the contact constraints. If two feet are in contact with the ground ($n_c=2; \mathbf{J}_{ab}^T\in\mathbb{R}^{6\times 6}$), there exists exactly one solution to (\ref{eq:base_tau}). The QP, in that case, gives us an estimate of the solution that satisfies the contact constraints and is closest to the analytical solution of (\ref{eq:base_tau}). If less than two feet are in contact with the ground $(n_c<2;\mathbf{J}_{ab}^T\in\mathbb{R}^{6\times (6-m)};m\in\{3,6\})$, there exists no exact solution for (\ref{eq:base_tau}), which implies the solution of the QP will satisfy the original equation in a least-squared error sense with a non-zero residual while satisfying the contact constraints. Note that for $n_c\ge 2$, $rank(\mathbf{J}_{ab}^T)=6$, while for $n_c<2$, $rank(\mathbf{J}_{ab}^T)\in\{3, 0\}$. 
 
In this work, we assume that the Jacobian matrices involved in the computations of the controller never become row-rank deficient or singular. The row rank sufficiency required for $\mathbf{J}_{ab}^T$ is ensured by the choice of gait, and non-singularity of $\mathbf{J}_{aa}^T$ by choosing the appropriate step length in the motion planner generating the base and foot trajectories and is explained in Section \ref{sec:p&e}.

\section{Optimization Methodology}
\label{sec:optim}
To solve the optimization problem (\ref{eq:optimization}), we reformulate the objective of the optimization problem in the standard quadratic form, as shown below
\begin{equation}
\begin{aligned}
    \arg \min_\mathbf{f}\ &\mathbf{f}^T\mathbf{Q}\mathbf{f}+\mathbf{P}\mathbf{f}+\mathbf{R}\\
           &\text{s.t.  } \mathbf{g}(\mathbf{f}) \le \mathbf{0},
    \label{eq:std_form}
\end{aligned}
\end{equation}
with the following definitions, 
\begin{align*}
    \mathbf{f} &= \mathbf{F}_c,\\
    \mathbf{Q} &\triangleq \mathbf{J}_{ab}\mathbf{S}_1\mathbf{J}_{ab}^{T}+\mathbf{W}+\mathbf{V},\\
    \mathbf{P} &\triangleq-2(\boldsymbol{\tau}_b^T\mathbf{S}_1\mathbf{J}_{ab}^T+\mathbf{f}^{*T}_{prev}\mathbf{V}),\\
    \mathbf{R} &\triangleq\boldsymbol{\tau_b}^{T}\mathbf{S}_1\boldsymbol{\tau_b}+\mathbf{f}^{*T}_{prev}\mathbf{V}\mathbf{f}^*_{prev},\\
    \mathcal{J} &\triangleq \mathbf{f}^{T}\mathbf{Q}\mathbf{f}+\mathbf{P}\mathbf{f}+\mathbf{R},
\end{align*}
where $\mathbf{f}_{prev}^*$ is the solution of the problem from the previous solution. The constraint vector, $\mathbf{g}(\mathbf{f})$, consists of the unilateral normal force and the friction cone constraints. The friction cone constraint makes $\mathbf{g}(\mathbf{f})$ nonlinear. This makes the optimization problem in (\ref{eq:std_form}) difficult to solve. The general approach in the existing literature is to linearize this constraint by approximating the friction cone by a friction pyramid. This makes it possible to use off-the-shelf QP solvers like qpOASES, OSQP, etc. Since we have a quadratic cost with a completely geometric constraint, i.e., the contact force should remain inside the friction cone, we propose an algorithm to solve this without linearizing the constraints named Geometric Projected Gradient Descent (GPGD). In this approach, we use gradient descent to find the solution to the optimization problem and manually enforce the constraint by projecting the resulting forces on each leg that is in contact with the ground onto the friction cone. The proposed algorithm solves the problem with the original nonlinear constraints without any simplification. 

The gradient and the Hessian used in a single step of the optimization algorithm are
\begin{align*}
    \nabla\mathcal{J}&=(\mathbf{Q} + \mathbf{Q}^T)\mathbf{f}+\mathbf{P}^T\\
    &= 2\mathbf{Q}\mathbf{f}+\mathbf{P}^T\ \ (\text{since } \mathbf{Q}=\mathbf{Q}^T), \\
    \nabla^2\mathcal{J}&=2\mathbf{Q}.
\end{align*}
\begin{figure}
    \centering
    \includegraphics[scale=0.7]{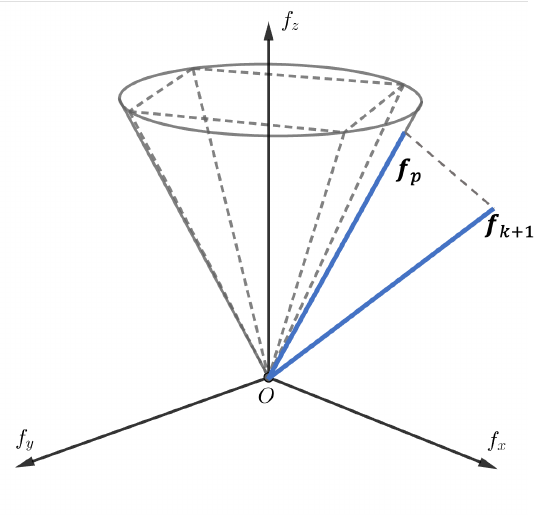}
    \caption{Projection of the iterative solution on the friction cone.}
    \label{fig:friction_projection}
\end{figure}
The projection step in the proposed algorithm is inspired by the algorithm given in (\cite{friction_projection}). It is simplified and shown in Fig. \ref{fig:friction_projection}.  
The proposed algorithm is given in the block marked \textbf{Algorithm \ref{alg:gpgd}}. In the algorithm, $\mathbf{f}_{n,i}$ denotes the contact force at iteration $n$ for $i$-th foot in contact with the ground.

\begin{algorithm}
\caption{Geometric Projected Gradient Descent}
\label{alg:gpgd}
\begin{algorithmic}[1]
\Function{Gpgd}{ }
    \State Initialize: $n \gets 0$, $\mathbf{f}_n \gets \mathbf{F}_{c_k}^*$, $\mathbf{f}_p \gets \mathbf{0}$
    \While{$n < \text{max\_iters}$}
        \State $\nabla\mathcal{J} \gets 2\mathbf{Q}\mathbf{f}_n + \mathbf{P}^T$  \Comment{update gradient}
        \State $\mathbf{f}_{n+1} \gets \mathbf{f}_n - (\nabla^2\mathcal{J})^{-1}\nabla\mathcal{J}$ \Comment{GD step}
        \For{$i \in \{1,\ldots,n_c\}$}
            \State $f_{n+1,i}^z \gets \max\{f_{n+1,i}^z, f_{\min}^z\}$ 
            \State $f_{n+1,i}^z \gets \min\{f_{n+1,i}^z, f_{\max}^z\}$ \Comment{unilaterality}
            \State $f_t \gets \|\mathbf{f}_{n+1,i}^{x,y}\|$
            \State $f_z \gets f_{n+1,i}^z$
            \If{$f_t \leq \mu f_z$} \Comment{inside friction cone}
                \State $\mathbf{f}_{p,i} \gets \mathbf{f}_{n+1,i}$
            \ElsIf{$f_t \leq -\frac{f_z}{\mu}$} \Comment{inside dual cone}
                \State $\mathbf{f}_{p,i} \gets \mathbf{0}$
            \Else \Comment{project on friction cone surface}
                \State $f_{p,i}^z \gets \frac{\mu f_t + f_z}{\mu^2+1}$
                \State $f_{p,i}^z \gets \min\{\max\{f_{p,i}^z, f_{\min}^z\}, f_{\max}^z\}$
                \State $\mathbf{f}_{p,i}^{x,y} \gets \mu f_{p,i}^z \frac{\mathbf{f}_{n+1,i}^{x,y}}{f_t}$
            \EndIf
        \EndFor
        \If{$\|\mathbf{f}_p - \mathbf{f}_n\|_{2} < 10^{-2}$} \Comment{check convergence}
            \State \Return $\mathbf{f}_p$  \Comment{return converged solution}
        \EndIf
        \State $n \gets n + 1$ \Comment{update iteration counter}
    \EndWhile
    \State \Return $\mathbf{f}_p$ \Comment{return sub-optimal solution}
\EndFunction
\end{algorithmic}
\end{algorithm}

\section{Planning and Estimation}
\label{sec:p&e}
\subsection{Planning}
The planner in the planning-estimation-control (PEC) loop computes the desired positions of the robot base and feet. The input to the planner is the commanded velocity, $^{\mathcal{B}}_{}\mathbf{v}_{cmd}=\begin{bmatrix}
    v^x&v^y&\omega^z
\end{bmatrix}^T$, generated by scaling the joystick commands with appropriate velocity limits. This is used to compute the desired translational base motion along $X^{\mathcal{G}}$ and $Y^{\mathcal{G}}$ axes, rotational base motion about $Z^{\mathcal{G}}$ axis, and the stepping location for the swing legs at each time-step $k$ of the PEC loop. 

\subsubsection{Base motion planning}
The required linear acceleration of the base in frame $\mathcal{B}$ is given by
\begin{equation}
    ^{\mathcal{B}}_{}\ddot{\mathbf{r}}_{b_{k}}^{x,y} = K_b(^{\mathcal{B}}_{}\mathbf{v}_{cmd,k}^{x,y} -\; ^{\mathcal{B}}\dot{\mathbf{r}}_{b_{d,k-1}}^{x,y}),
\end{equation}
where $\dot{\mathbf{r}}_{b_{d,k-1}}^{x,y}$ is the desired base velocity at the previous time-step. The required base linear acceleration is saturated using the maximum allowed acceleration $\mathbf{a}_{\text{max}}$ to get the desired acceleration and then used to update the desired base velocity $\dot{\mathbf{r}}_{b_{d,k}}^{x,y}$ and position $\mathbf{r}_{b_{d,k}}^{x,y}$ as shown in (\ref{eq:ref_base}).
\begin{equation}
\begin{aligned}
    ^{\mathcal{B}}_{}\ddot{\mathbf{r}}_{b_{d, k}}^{x,y} &= \min\{\max\{^{\mathcal{B}}_{}\ddot{\mathbf{r}}_{b_{k}}^{x,y}, -\;^{\mathcal{B}}\mathbf{a}_{\text{max}}^{x,y}\}, \;^{\mathcal{B}}\mathbf{a}_{\text{max}}^{x,y}\}, \\
    ^{\mathcal{G}}\dot{\mathbf{r}}_{b_{d,k}}^{x,y}&=\; ^{\mathcal{G}}\dot{\mathbf{r}}_{b_{d,k-1}}^{x,y} +\; ^{\mathcal{G}}_{\mathcal{B}}\mathbf{R}\;^{\mathcal{B}}_{}\ddot{\mathbf{r}}_{b_{d, k}}^{x,y}\Delta t,\\
    ^{\mathcal{G}}\mathbf{r}_{b_{d,k}}^{x,y}&=\;^{\mathcal{G}}\mathbf{r}_{b_{d,k-1}}^{x,y}+\;^{\mathcal{G}}\dot{\mathbf{r}}_{b_{d,k}}^{x,y}\Delta t+\frac{1}{2}\;^{\mathcal{G}}\ddot{\mathbf{r}}_{b_{d,k}}^{x,y}\Delta t^2,
    \label{eq:ref_base}
\end{aligned}
\end{equation}
where $^{\mathcal{G}}_{\mathcal{B}}\mathbf{R}$ is the rotation matrix representing the transformation from the local to the global frame, and $\Delta t$ is the difference between the current time and the time at the end of the previous iteration. The maximum velocity values are chosen based on the base height and the gait frequency. The base height imposes a kinematic constraint on the maximum step length of the robot. For acceleration limits, we multiply $F^z_{c,\text{max}}$ by the friction coefficient to get the maximum horizontal force, then divide it by the total mass of the robot to get the maximum possible acceleration. $K_b$ is the gain chosen based on the desired first-order response of the velocity. Similarly, the desired yaw motion is calculated as
\begin{align*}
    \ddot{\theta}_{d,k}^z&=\min\{\max\{K_{\theta}(\omega^z_k - \dot{\theta}_{d,k-1}^z), -\alpha_{\text{max}}^z\}, \alpha_{\text{max}}^z\},\\
    \dot{\theta}_{d,k}^z&=\min\{\max\{\dot{\theta}_{d,k-1}^z + \ddot{\theta}_{d,k}^z\Delta t, -\omega_{\text{max}}^z\}, \omega_{\text{max}}^z\},\\
    \theta_{d,k}^z&=\theta_{d,k-1}^z+\dot{\theta}_{d,k}^z\Delta t+\frac{1}{2}\ddot{\theta}_{d,k}^z\Delta t^2,
\end{align*}
where $\theta^z$ is the rotation angle about the $Z^\mathcal{G}$ axis, $\alpha_{\text{max}}^z$ and $\omega_{\text{max}}^z$ are the fixed maximum angular acceleration and velocity about the $Z^\mathcal{G}$ axis, respectively. This desired angle is converted to the specific orientation representation being used before being passed to the controller for tracking.

\subsubsection{Foothold planning}
To plan the footholds, a desired gait is predefined that specifies when each leg will be in the swing phase or the stance phase. Inspired by the Hildebrand gait parameters (\cite{hildebrand}), we define a gait using four variables: 
\begin{enumerate}[label=\alph*)]
    \item gait period $t_g$,
    \item duty factor $\phi_{st,i}$,
    \item phase offset $\psi_{o,i}$, and
    \item step height $h_z$.
\end{enumerate}
The variables with subscript $i$ are defined for each leg individually. The duty factor $\phi_{st,i}$ is the fraction of the normalized gait period after which the contact state of the $i$-th leg changes from stance to swing. The phase offset $\psi_{o,i}$ is the normalized timing offset of each leg. $\psi_{o,i}$ along with $\phi_{st,i}$ define the type of gait the robot will execute (walk, trot, bound, etc.). The desired contact state $s_{d_{i,k}}$ is calculated as
\begin{align*}
    \phi_{ph,i}&=\mod(t/t_g+\psi_{o,i}, 1),\\
    s_{d_{i,k}}&=\begin{cases}
      1 & \text{if $\frac{\phi_{ph,i}}{\phi_{st}} < 1$}\\
      0 & \text{otherwise}
    \end{cases},
\end{align*}
where $s_{d_{i,k}} = 1$ corresponds to stance and $s_{d_{i,k}} = 0$ to swing. For each foot in swing, its landing position $\mathbf{r}_{p_{i,d,k}}^{x,y}$ is calculated using the following foot placement heuristic
\begin{equation}
\begin{aligned}
     ^{\mathcal{G}}\mathbf{r}_{p_{i,d,k}}^{x,y} &= \;^{\mathcal{G}}\mathbf{r}_{h_{i,d,k}}^{x,y} + \;^{\mathcal{G}}\mathbf{d}_{i}^{x,y}, \\
     ^{\mathcal{G}}\mathbf{d}_{i}^{x,y} &= \frac{1}{2}\;^{\mathcal{G}}\dot{\mathbf{r}}_{h_{i,d,k}}^{x,y}\phi_{st,i}t_g,
\end{aligned}
\end{equation}
where $\mathbf{r}_{h_{i,d,k}}^{x,y}$ is the position of the $i$-th hip, and $\mathbf{d}_{i}^{x,y}$ is the position of the desired foothold from the current hip position, calculated using the Raibert heuristic (\cite{raibert_book}). The norm of $\mathbf{d}_{i}^{x,y}$ is the step length for the $i$-th leg in the horizontal plane. For flat ground, $^{\mathcal{G}}\mathbf{d}_{i}^{z} = 0$ as the terrain height is uniform. The position and velocity of the $i$-th hip is given by
\begin{equation}
\begin{aligned}
     ^{\mathcal{G}}\mathbf{r}_{h_i} &= \;^{\mathcal{G}}\mathbf{r}_b + \;^{\mathcal{G}}_{\mathcal{B}}\mathbf{R}\;^{\mathcal{B}}\mathbf{r}_{h_i/b},\\
    ^{\mathcal{G}}\dot{\mathbf{r}}_{h_i} &= \;^{\mathcal{G}}\dot{\mathbf{r}}_{b} + \;^{\mathcal{G}}\boldsymbol{\omega}_{b}\times\;^{\mathcal{G}}_{\mathcal{B}}\mathbf{R}\;^{\mathcal{B}}\mathbf{r}_{h_i/b},
\end{aligned}
\end{equation}
where $\mathbf{r}_{h_i/b}$ is the vector from the COM to the $i$-th hip and $\boldsymbol{\omega}_{b}$ is the angular velocity of the base. This swing foot trajectory is composed of sinusoidal functions.

The step length is limited by the geometry of the robot, specifically, the hip height $r_{h_i}^z$, and maximum leg extension $l_e$. The maximum allowable step length can be calculated using
\begin{equation}
    d_{i,\text{max}} = \sqrt{\beta l_e^2-(r_{h_i}^z)^2},
\end{equation}
where the factor $\beta\in \left[\left(\frac{r_{h_i}^z}{l_e}\right)^2,1\right)$ is used to avoid a singular configuration for the leg. When $\beta$ is unity, $d_{i,max}$ is the maximum step length for which the $i$-th leg reaches a singular position. We modify the step location $\mathbf{d}_i^{x,y}$ and base height $r_{b_{d,k}}^z$ as explained in \textbf{Algorithm \ref{alg:ls_calc}}. This ensures that the desired footholds are always reachable while staying away from the singular configuration. Moreover, avoiding near-singular configurations of legs is favorable for force generation capability as this metric is directly related to the singular values of the matrix $\mathbf{J}_{aa}^T$.

\begin{algorithm}
\caption{Adaptive Step-Length and Base Height}
\label{alg:ls_calc}
\begin{algorithmic}[1]
\Function{AdaptStepLength}{ }
    \State $d_{\text{limit}} \gets \textbf{false}$
    \For{$i \in \{1,\ldots,4\}$}
        \State $\mathbf{d}_i^{x,y} \gets \frac{1}{2}\dot{\mathbf{r}}_{h_{i,d,k}}^{x,y}\phi_{st,i}t_g$
        \State $d_i \gets \|\mathbf{d}_i^{x,y}\|_{2}$
        \State $d_{i,\text{max}} \gets \sqrt{\beta l_e^2 - (r_{h_i}^z)^2}$
        \If{$d_i \geq d_{i,\text{max}}$}
            \State $d_{\text{limit}} \gets \textbf{true}$
            \State $\mathbf{d}_i^{x,y} \gets \frac{\mathbf{d}_i^{x,y}}{d_i}d_{i,\text{max}}$ \Comment{adapt step-length}
        \EndIf
    \EndFor
    \If{$d_{\text{limit}}$ is $\textbf{true}$} \Comment{adapt base height}
        \State $r_{b_{d,k}}^z \gets r_{b_{d,k}}^z - \alpha_- \cdot \Delta t$
    \Else
        \State $r_{b_{d,k}}^z \gets r_{b_{d,k}}^z + \alpha_+ \cdot \Delta t$
    \EndIf
    \State $r_{b_{d,k}}^z \gets \min\{\max\{r_{b_{d,k}}^z, r^{z}_{b,\text{min}}\}, r^{z}_{b,\text{max}}\}$
    \State \Return $r_{b_{d,k}}^z$
\EndFunction
\end{algorithmic}
\end{algorithm}

In \textbf{Algorithm \ref{alg:ls_calc}}, the factors $\alpha_-$ and $\alpha_+$ control the rate of reference base height decrement and increment, respectively. 

\begin{figure}[H]
    \centering
    \includegraphics[scale=0.5]{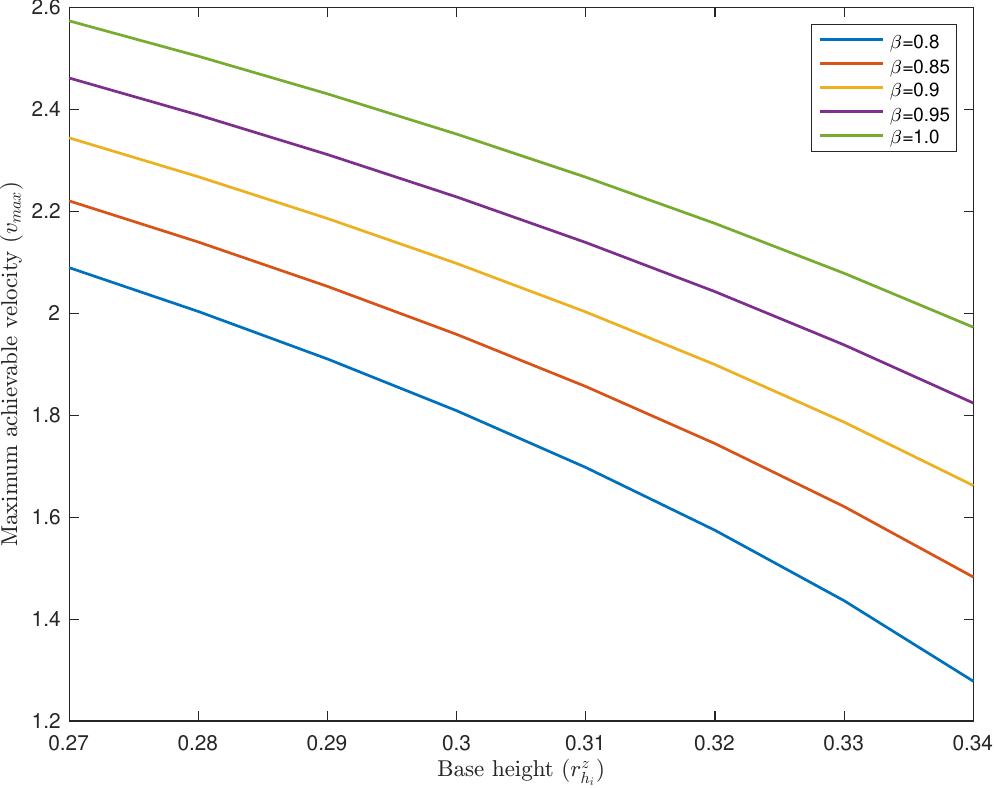}
    \caption{Variation of maximum achievable velocity with base height for different values of $\beta$.}
    \label{fig:rhz_vs_vmax}
\end{figure}

The choice of $\beta$, $\phi_{st}$, and $t_g$ determine the maximum achievable base velocity using the described planning technique. For a chosen value of $\beta$, the theoretical maximum velocity ($v_{max}$) that the robot can have is:
\begin{equation}
    \begin{aligned}
        v_{max}=\frac{2\sqrt{\beta l_e^2-(r_{h_i}^z)^2}}{\phi_{st}t_g}.
    \end{aligned}
\end{equation}
In the above equation, the maximum velocity for a chosen gait (fixed $\phi_{st}$ and $t_g$), is dependent on the base height and the parameter $\beta$. This relation is shown in Fig. \ref{fig:rhz_vs_vmax}. For the figure, $\phi_{st}=0.5$, $t_g=0.5$, and $l_e=0.42$ were chosen.
This figure illustrates how base height adjustment discussed in \textbf{Algorithm \ref{alg:ls_calc}}, triggered when the step length limit is reached, enables the robot to continue motion at the desired speed while avoiding singularity.

Note that in this analysis, the collision between any two feet (or, legs) is not considered, but it provides an upper bound on the expected maximum velocity of the robot. This is useful because even when other constraints (e.g., box constraints) are placed on step length, the overall variation of the maximum velocity with the hip height ($r_{h_i}^z$) and $\beta$ remains the same.

\subsection{State estimation}
The full state estimation pipeline consists of interdependent base pose and contact state estimators.

\subsubsection{Base pose estimation}
The robot base pose (position and orientation) and its derivative (linear and angular velocity) are estimated using the IMU and joint encoder. The framework used here is based on the formulation proposed in (\cite{bloesch}) and (\cite{cheetah3}). The acceleration and gyroscope data from IMU are used for the prediction model of the Kalman filter. The orientation data is taken directly from the IMU. For the correction step of the estimator, a no-slip condition is assumed on the stance feet to update the robot base linear position and velocity.

\subsubsection{Contact estimation}
For contact estimation, the measured torques from the motors are used to estimate the contact forces on the feet first. The EOM described in (\ref{eq:eom}) is used for estimating foot forces as
\begin{align*}
    \hat{\mathbf{F}}_c=-(\mathbf{J}_{c}^T(\hat{\mathbf{q}}))^{\dagger}(\mathbf{S}^T\boldsymbol{\hat{\tau}} - \mathbf{M}(\hat{\mathbf{q}})\hat{\ddot{\mathbf{q}}} - \boldsymbol{\eta}(\hat{\mathbf{q}}, \hat{\dot{\mathbf{q}}})),
\end{align*}
where $(\hat{.})$ represents the estimated quantities or filtered sensor data. The estimated generalized coordinate vector $\hat{\mathbf{q}}$ and generalized velocity vector $\hat{\dot{\mathbf{q}}}$ are constructed by combining the output of the base pose estimator and filtered sensor values. The estimated generalized acceleration vector, $\hat{\ddot{\mathbf{q}}}$, is calculated by using a filtered finite-difference (\cite{savitzky_golay}) of the estimated generalized velocity vector $\hat{\dot{\mathbf{q}}}$. The estimated contact force, $\hat{\mathbf{F}}_c$, coupled with the estimated feet height and reference contact phase as per the gait scheduler, are used to estimate the contact probabilities $P_{c_{i,k}}$ using a Kalman filter-based approach described in (\cite{contact_estimation}). The estimated contact state $s_{i,k}$ is computed using hysteresis thresholding on $P_{c_{i,k}}$ to make the contact state prediction more stable.

\begin{align*}
    s_{i,k} = \begin{cases}
      1 & \text{if $P_{c_{i,k}} \ge 0.6$}\\
      0 & \text{if $P_{c_{i,k}} \le 0.4$}\\
      s_{i,k-1} & \text{else}
    \end{cases},
\end{align*}
where $P_{c_{i,k}}$ is the contact probability and $s_{i,k}$ is the contact state for the leg $i$ for the current time-step $k$.

\section{Experiments and Results}
\label{sec:experiments}
\begin{figure}
    \centering
    \includegraphics[scale=0.34]{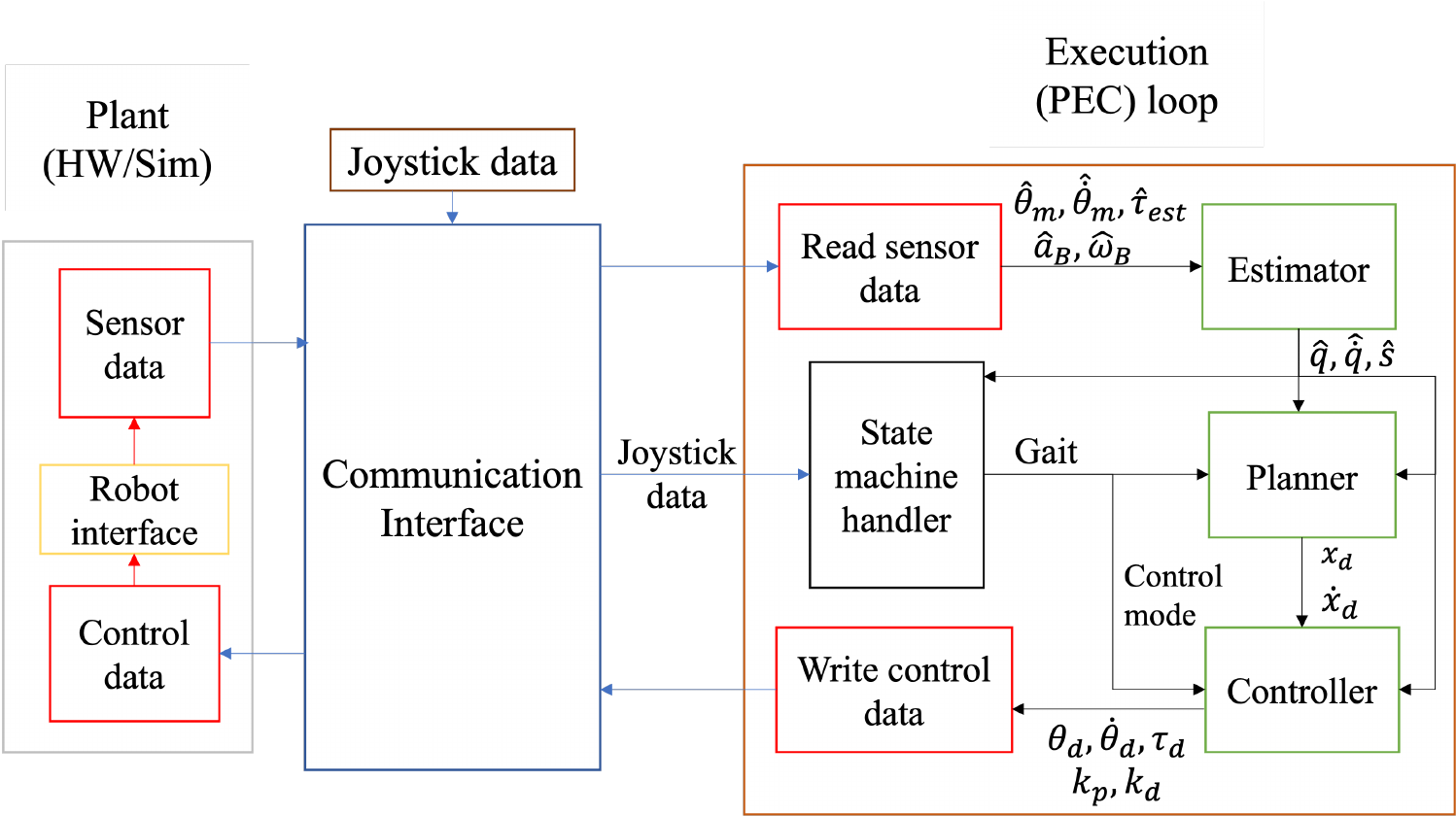}
    \caption{Software architecture.}
    \label{fig:sw_arch}
\end{figure}

\begin{figure}
    \centering
    \includegraphics[scale=0.34]{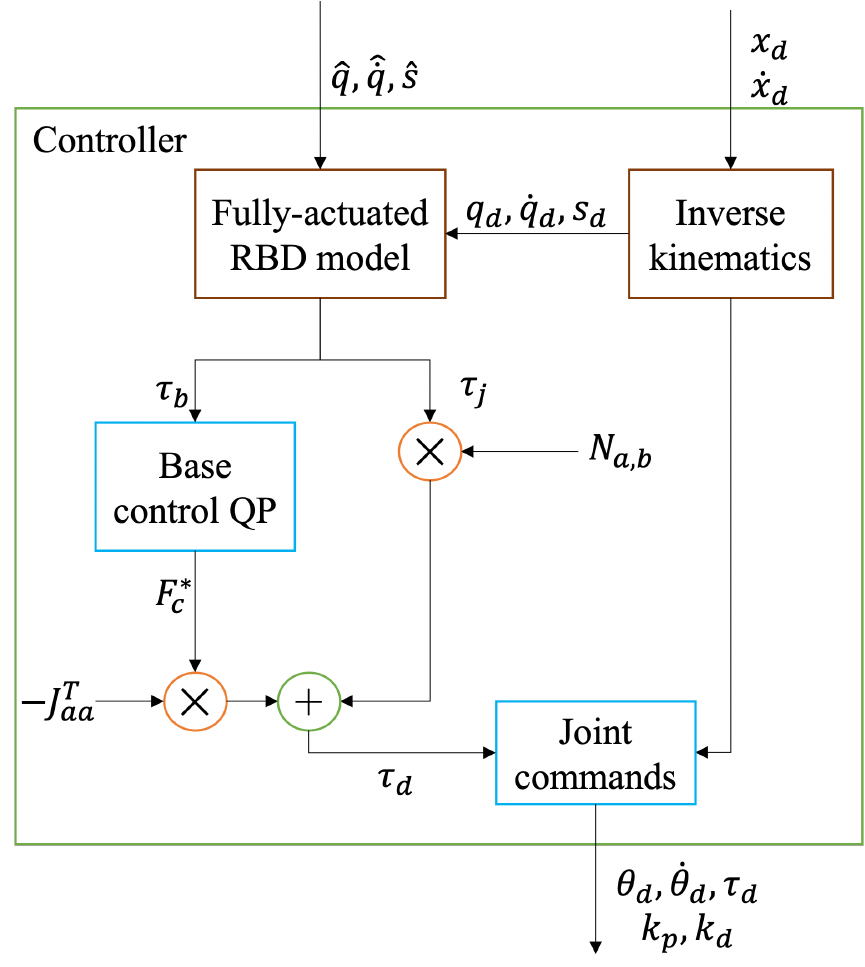}
    \caption{Controller architecture.}
    \label{fig:ctrl_arch}
\end{figure}

The proposed controller is tested and validated on 
Go2 robot from Unitree Robotics. The simulation experiments were conducted in the MuJoCo (\cite{mujoco}) simulator on an Apple M1 Pro ARM64 CPU. The software architecture used for the experiments is shown in Fig. \ref{fig:sw_arch}. Inside the communication interface, CycloneDDS (\cite{cyclonedds}) was chosen.
The controller block in Fig. \ref{fig:sw_arch} is expanded in Fig. \ref{fig:ctrl_arch} to show further details about how joint commands being sent to the motor controller are generated.
The gains $\mathbf{K}_p$ and $\mathbf{K}_d$ can be calculated by formulating the control problem for each DOF as a linear quadratic regulator (\cite{lqr}). The advantage of calculating the gains using this approach is that they are guaranteed to maintain the stability of the decoupled DOFs. For this, every component of (\ref{eq:lin_dyn}) can be written in the state space form as
\begin{align*}
    \dot{\mathbf{x}} = \mathbf{A}\mathbf{x}+\mathbf{B}u
\end{align*}
where $\mathbf{x} = [q_i\ \dot{q}_i]^T$, $u = \ddot{q}_{cmd, i}$, and
\begin{align*}
    \mathbf{A} = \begin{bmatrix}
        0&1\\0&0
    \end{bmatrix}, \mathbf{B} = \begin{bmatrix}
        0\\1
    \end{bmatrix}.
\end{align*}
The infinite time LQR problem for this is defined as
\begin{align*}
    &\min_{u}\int_{0}^{\infty}\left(\mathbf{x}^T\mathbf{Q}\mathbf{x}+u^TRu\right)dt\\
    &s.t.\ \ \ \ \dot{\mathbf{x}} = \mathbf{A}\mathbf{x}+\mathbf{B}u.
\end{align*} 
The chosen state and control weights are: $\mathbf{Q} = diag([100, 1])$ and $R = 10^{-3}$ for each DOF of the decoupled dynamics.  
We modify the base translational gains in $x$ and $y$ axis to be $K_p=0$ and $K_d=10.4$. 
The relative weighting matrices in the optimization cost are chosen heuristically (but to achieve reasonable performance): $\mathbf{S}_1=diag(1,1,2,20,20,5)$, $\mathbf{W}=10^{-2}$, $\mathbf{V}=10^{-3}$.
\subsection{Simulation results}
In the simulation environment, rate limiting and latency are added to make it similar to the actual hardware system. The original update rate of the simulation is kept at 1000 Hz. We update the control and sensor data objects at 500 Hz. To introduce latency in the simulation, we first make a probabilistic map of the latency times in bins of 5ms. While updating the control commands, we first push the incoming data at the start of a queue. We then use the probability map of the system latency to calculate the index for the control command to be used from the queue at the current time step. 
\subsubsection{Locomotion performance}
\begin{figure}
    \centering
    \includegraphics[width=1.0\linewidth]{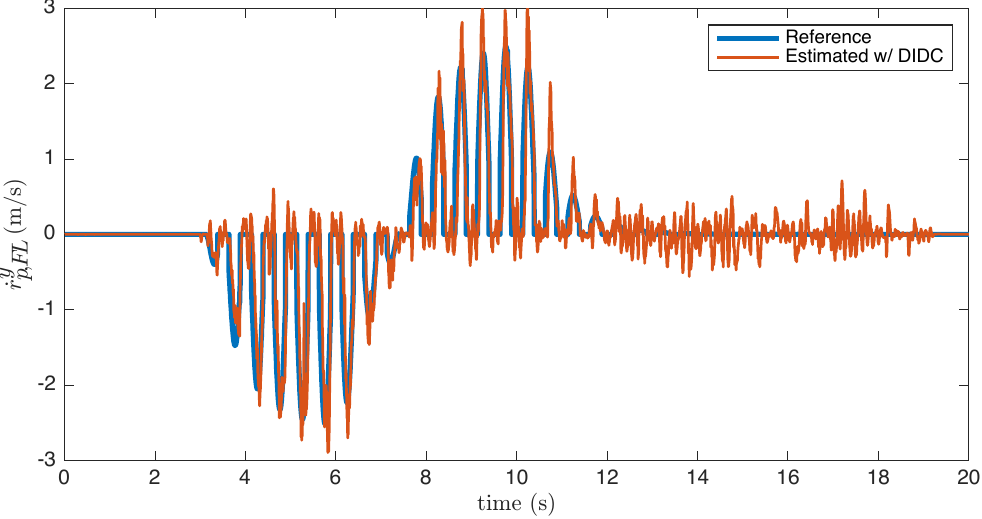}
    \includegraphics[width=1.0\linewidth]{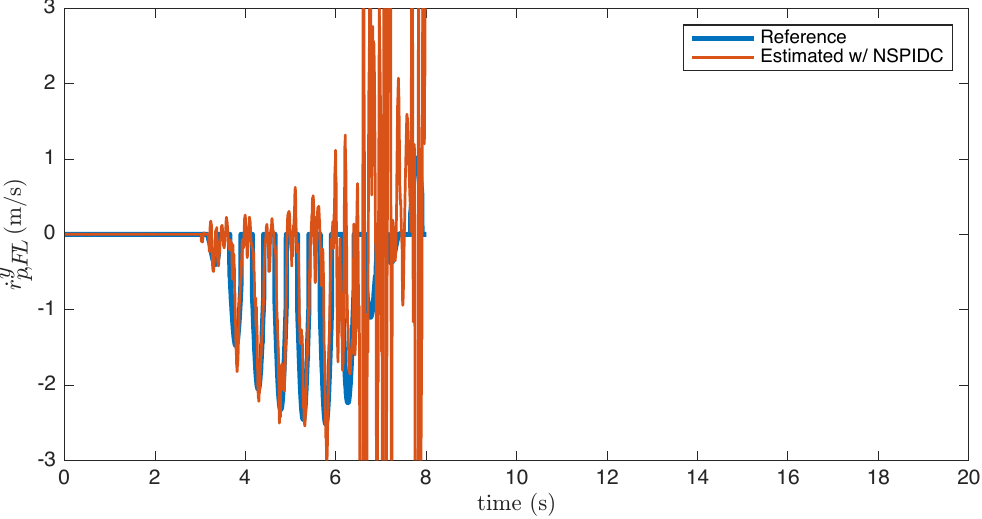}
    \caption{Estimated and reference $Y^{\mathcal{G}}$-axis velocity of the FL foot in simulation.}
    \label{fig:sim_feet_vel}
\end{figure}

\begin{figure}
    \centering
    \includegraphics[width=1.0\linewidth]{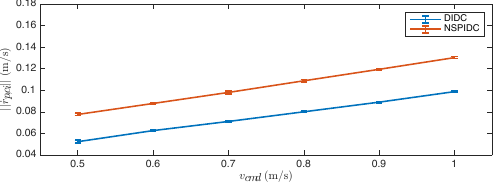}
    \caption{Variation of average foot slip with increasing commanded velocities in simulation.}
    \label{fig:foot_slip_stats}
\end{figure}

\begin{figure}
    \centering
    \includegraphics[width=1.0\linewidth]{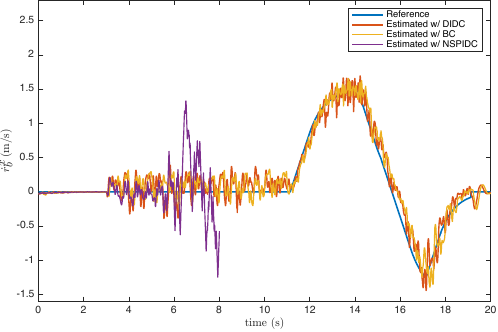}
    \includegraphics[width=1.0\linewidth]{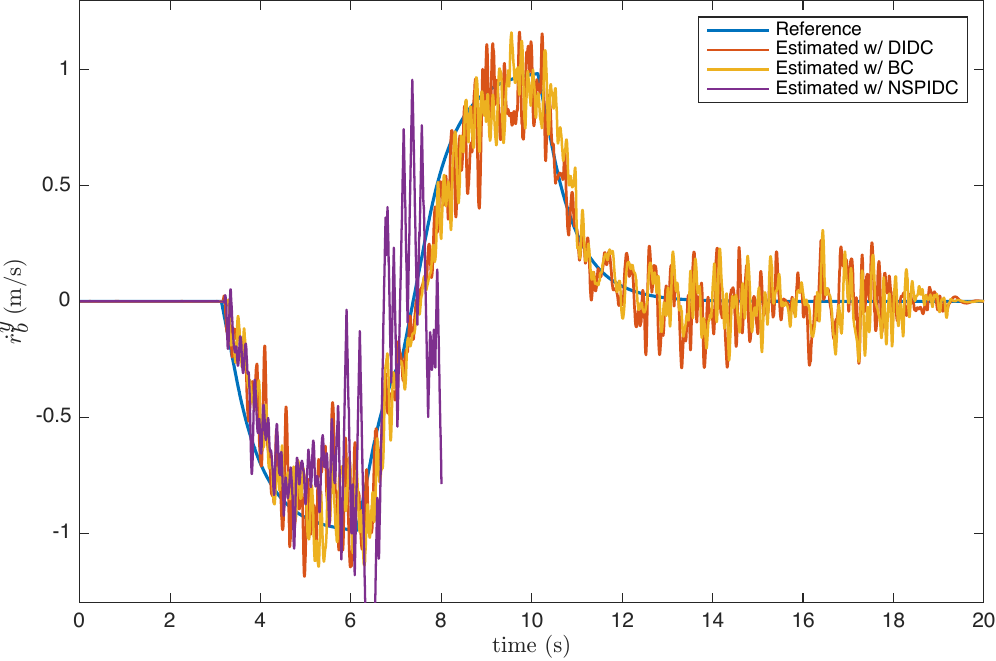}
    \caption{Estimated and reference velocity of the robot base in simulation.}
    \label{fig:sim_base_vel}
\end{figure}

The simulation experiment consists of the motion of the robot in $X^{\mathcal{G}}$ and $Y^{\mathcal{G}}$ directions with a velocity of $\mp1.0$ m/s along $Y^{\mathcal{G}}$ followed by $\pm1.5$ m/s along $X^{\mathcal{G}}$. Fig. \ref{fig:sim_feet_vel} shows feet velocity tracking for DIDC and NSPIDC, to show the effect of friction cone constraints.
In the experiment, foot slip occurs for NSPIDC when the robot starts moving at high velocity, causing the robot to fall. Hence, the data for collection for NSPIDC is stopped before the experiment can be completed. This is the reason for only partial plots for NSPIDC in figures of single runs. As discussed in Section II, this is due to friction cone constraint violation that is not enforced in the NSPIDC controller. To show the variation of foot slip with the NSPIDC controller, a different set of softer gains is chosen to make the NSPIDC controller work for moderately high velocities. Fig. \ref{fig:foot_slip_stats} shows the variation in average foot slip when they are in contact for different commanded velocities. With softer gains, the NSPIDC still has $\sim$32\% higher foot slip than DIDC. The controllers incorporate friction constraints (e.g., DIDC and BC) to track the desired velocity without failure due to foot slip. The base velocity tracking for the experiment with original LQR gains is shown in Fig. \ref{fig:sim_base_vel}.

\begin{figure}
    \centering
    \includegraphics[width=1.0\linewidth]{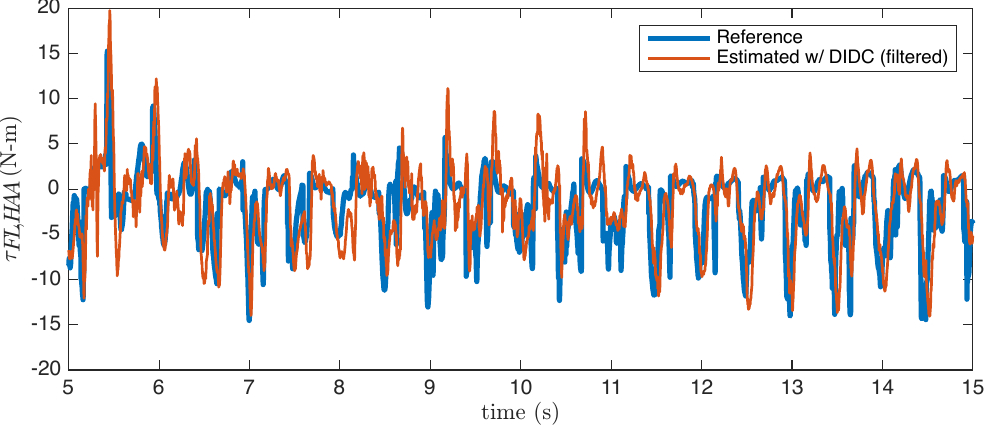}
    \includegraphics[width=1.0\linewidth]{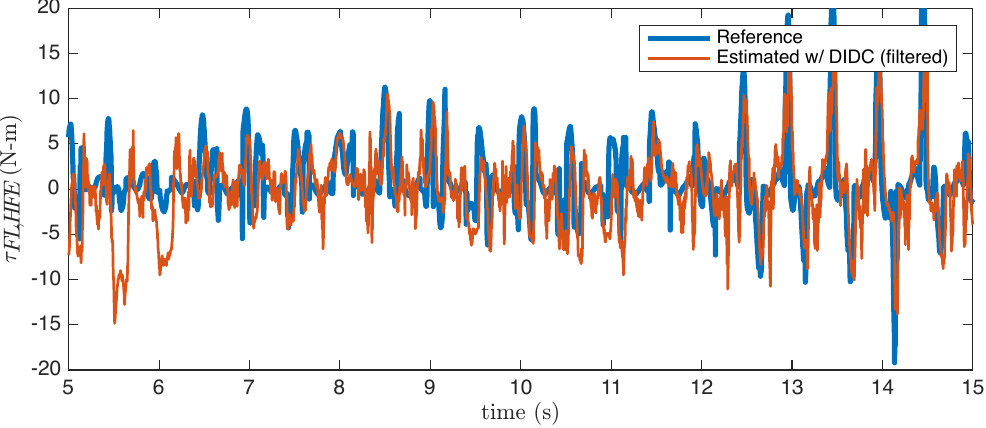}
    \includegraphics[width=1.0\linewidth]{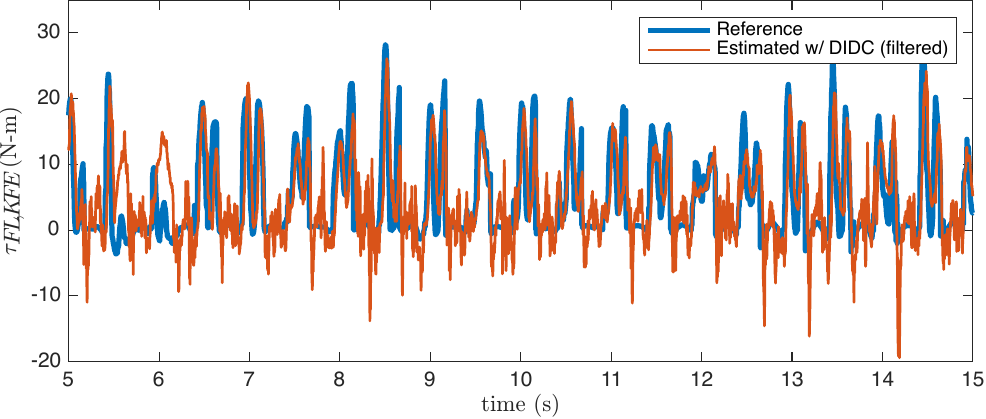}
    \caption{Estimated and reference torques of the FL leg joints in simulation.}
    \label{fig:sim_torque_tracking}
\end{figure}

\begin{figure}
    \centering
    \includegraphics[width=1.0\linewidth]{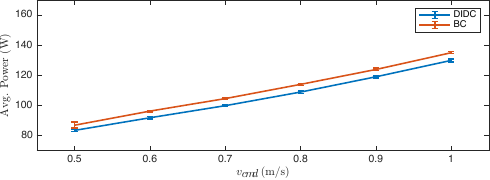}
    \caption{Comparison of average power required for increasing commanded velocity in simulation.}
    \label{fig:power_stats}
\end{figure}

\begin{figure}
    \centering
    \includegraphics[width=1.0\linewidth]{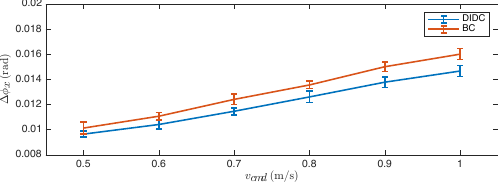}
    \includegraphics[width=1.0\linewidth]{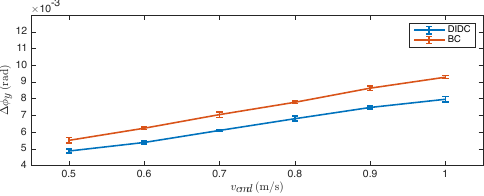}
    \caption{Error bar graph of the orientation (roll and pitch) error for different commanded velocities in simulation.}
    \label{fig:orientation_stats}
\end{figure}

The reference and the estimated torques for the motion are shown in Fig. \ref{fig:sim_torque_tracking}. The variation in average power consumed during the motion for different commanded velocities for  DIDC and BC is shown in Fig. \ref{fig:power_stats}. Despite both the controllers following the same routine, and hence having the same theoretical power requirement, DIDC uses lesser power ($\sim$5\%) in practice due to lesser feedback requirements compared to BC.
The variation of the orientation error for different commanded velocities in the experiment is shown in Fig. \ref{fig:orientation_stats}
The error bars in the figure indicate the difference of one standard deviation from the mean. Note that the controller feedback gains used for this experiment are the softer gains that work across all the controllers.
\subsubsection{Solver performance}
The proposed solver (GPGD) is compared against an off-the-shelf solver (qpOASES) to benchmark its performance. We record the average of the base wrench residual ($||\hat{\boldsymbol{\tau}}_b-\boldsymbol{\tau}_b||\triangleq\delta r$), time taken by the solver in milli-seconds ($t_{sol}$), and the total constraint violation ($||\boldsymbol{\delta}_{\mathbf{g}}||\triangleq\delta_g $), for 10 continuous runs with the same command schedule. Table \ref{table:solver_stats} contains the norm of the mean and standard deviation for all three quantities. To calculate constraint violation for GPGD, the friction cone violation is calculated,
\begin{equation}
    \begin{aligned}
        \boldsymbol{\delta}_{\mathbf{g},i}=\min\{\mu F_{c,i}^{z*}-\sqrt{(F_{c,i}^{x*})^2+(F_{c,i}^{y*})^2}, 0\},
    \end{aligned}
\end{equation}
whereas in qpOASES, the linear friction violation is calculated,
\begin{equation}
    \begin{aligned}
        \boldsymbol{\delta}_{\mathbf{g},i}=\min\{&\mu F_{c,i}^{z*}-F_{c,i}^{x*},\\
        &\mu F_{c,i}^{z*}+F_{c,i}^{x*},\\ 
        &\mu F_{c,i}^{z*}-F_{c,i}^{y*},\\
        &\mu F_{c,i}^{z*} + F_{c,i}^{y*},\\
        &0\}
    \end{aligned}.
\end{equation}
The comparison is shown in Table \ref{table:solver_stats} and contains the mean and standard deviation of the average for each run with 10 runs in total.
\begin{table}
\caption{Solver statistics comparison.}
\begin{tabularx}{1.0\linewidth} { 
   >{\centering\arraybackslash}X
   >{\centering\arraybackslash}p{1.5cm}
   >{\centering\arraybackslash}p{1.5cm} 
   >{\centering\arraybackslash}p{2.5cm} }
 \toprule
 \textbf{Solver} & \textbf{Residual} ($\delta r$) & \textbf{Time} (ms) ($t_{sol}$) & \textbf{Constraint Violation} ($\delta_g$) \\
 \hline
 \texttt{GPGD}  & $1.1\mathrm{e}0\pm4.5\mathrm{e}{-2}$ & $9.3\mathrm{e}{-3}\pm1.7\mathrm{e}{-4}$ & $-6.1\mathrm{e}{-3}\pm8.4\mathrm{e}{-4}$\\
\hline
\texttt{qpOASES}& $1.6\mathrm{e}0\pm5.4\mathrm{e}{-2}$ & $2.3\mathrm{e}{-2}\pm1.5\mathrm{e}{-3}$ & $-9.7\mathrm{e}{-1}\pm3.0\mathrm{e}{-2}$\\
\bottomrule
\end{tabularx}\\[8pt]
\label{table:solver_stats}
\end{table}
Compared to the general QP solver, the proposed solver has significantly lesser constraint violation ($\sim$$99\%$), is $\sim$$2.47$ times faster, and has lesser residual ($\sim$31\%) on average. Note that the timing comparison is for the naive implementation of the proposed algorithm against the highly optimized implementation of qpOASES. The iteration time for GPGD can be further reduced by improving its implementation.
\subsection{Hardware results}
\begin{figure}
    \centering
    \includegraphics[width=0.9\linewidth]{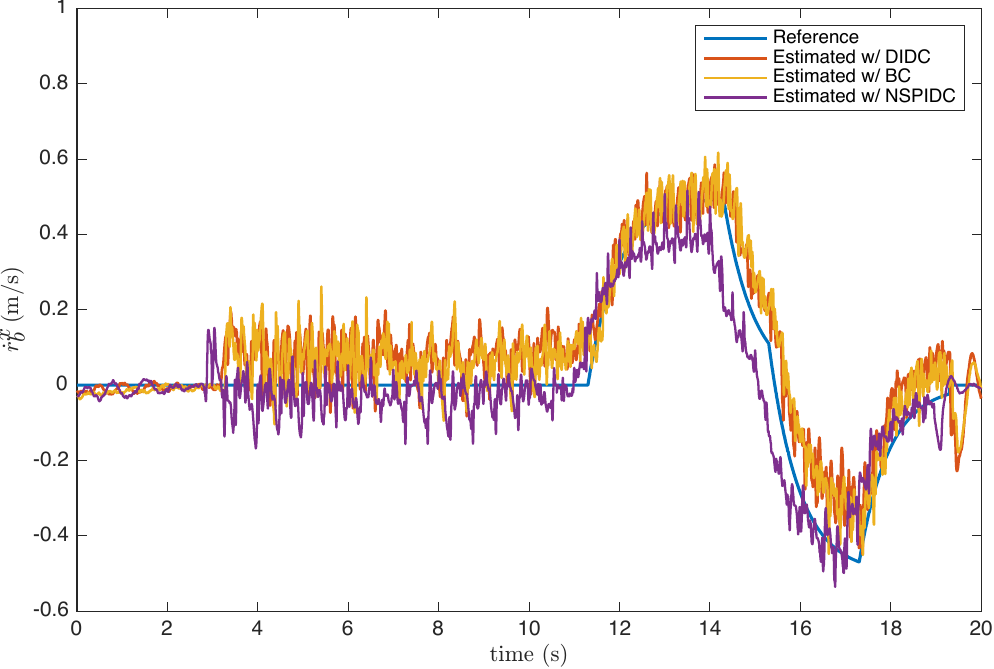}
    \includegraphics[width=0.9\linewidth]{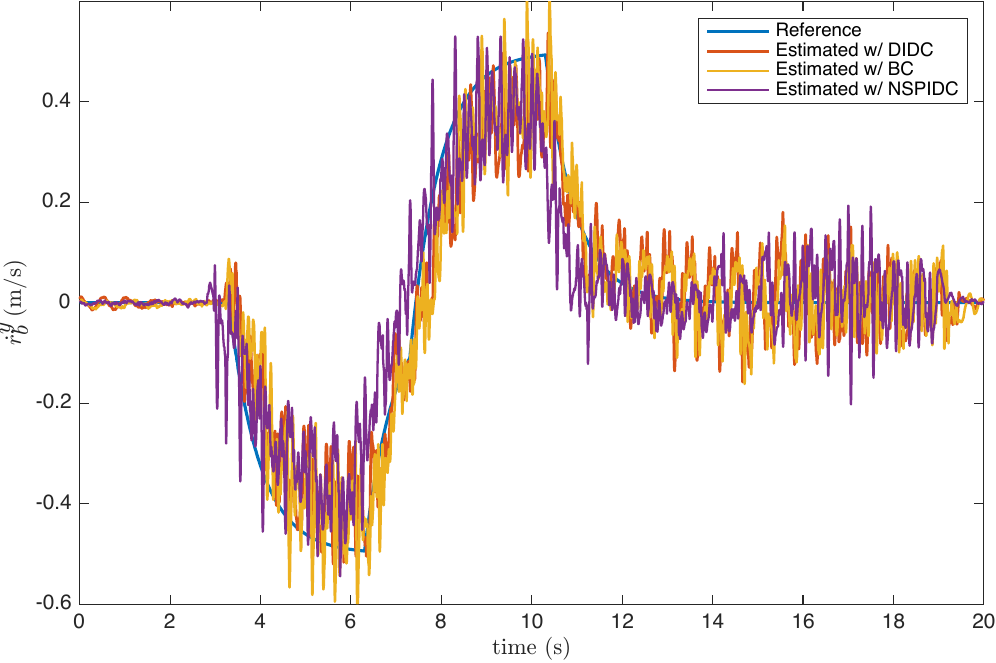}
    \caption{Estimated and reference velocity of the robot base on hardware.}
    \label{fig:velocity_tracking}
\end{figure}

\begin{figure}
    \centering
    \includegraphics[width=1.0\linewidth]{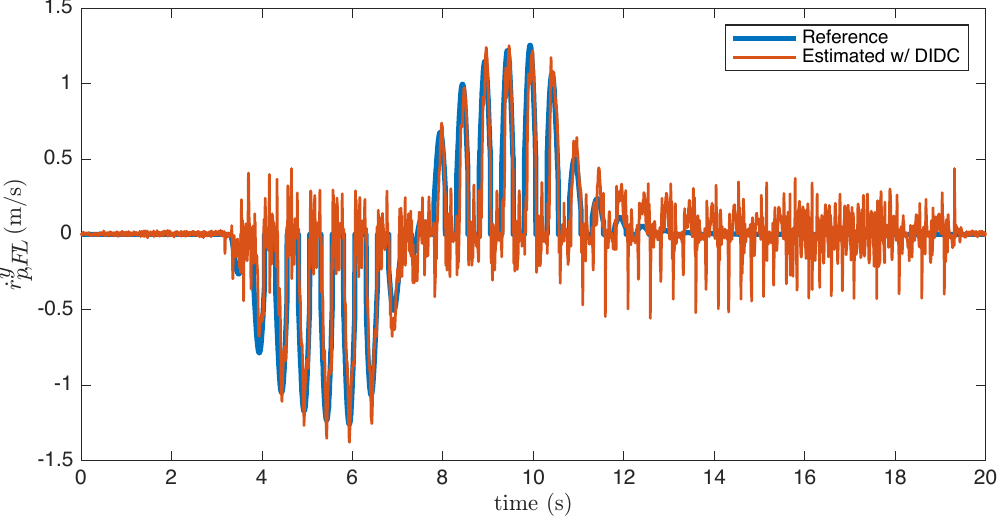}
    \includegraphics[width=1.0\linewidth]{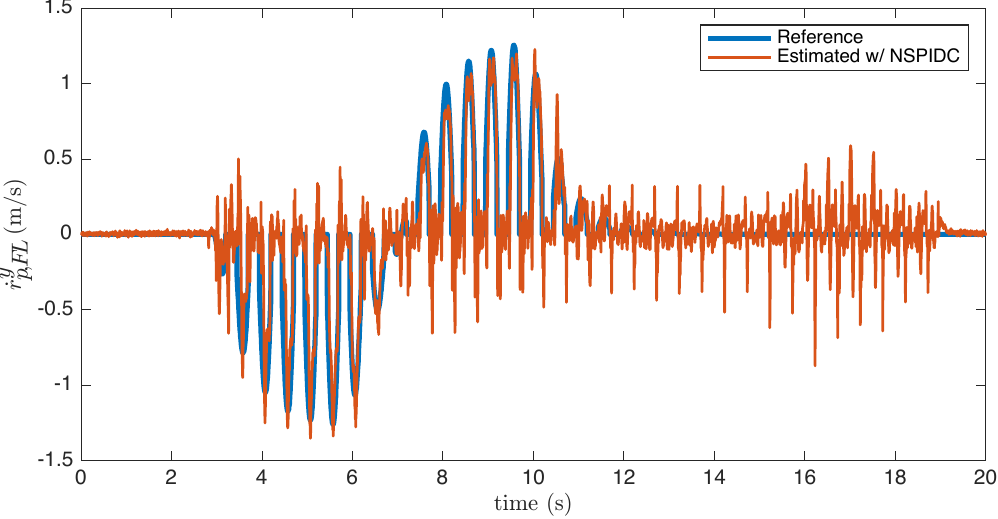}
    \caption{Estimated and reference $Y^{\mathcal{G}}$-axis velocity of the FL foot on hardware.}
    \label{fig:hw_feet_vel}
\end{figure}

\begin{figure}
    \centering
    \includegraphics[width=1.0\linewidth]{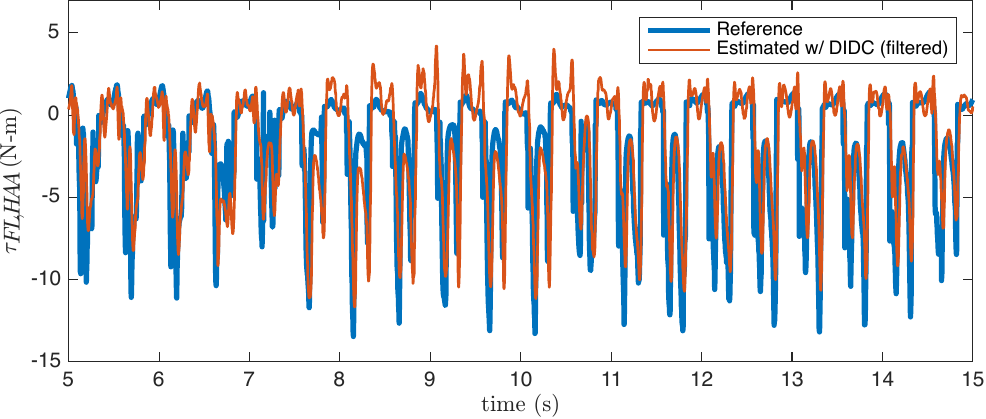}
    \includegraphics[width=1.0\linewidth]{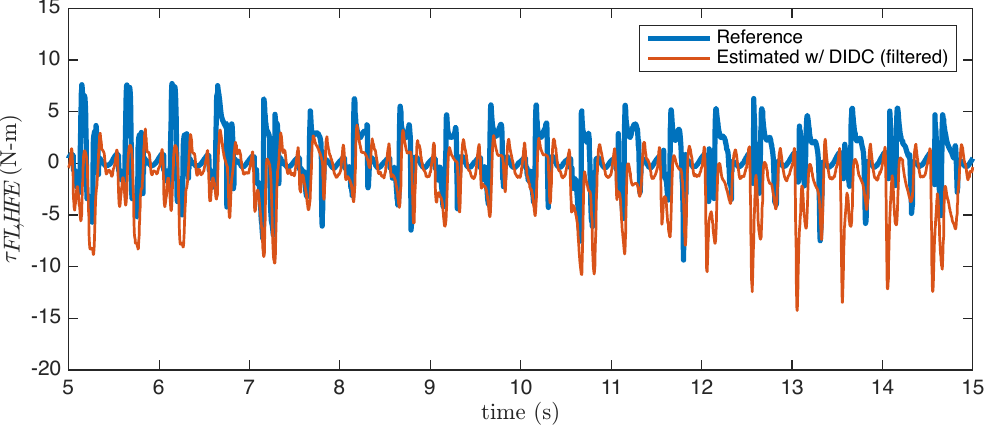}
    \includegraphics[width=1.0\linewidth]{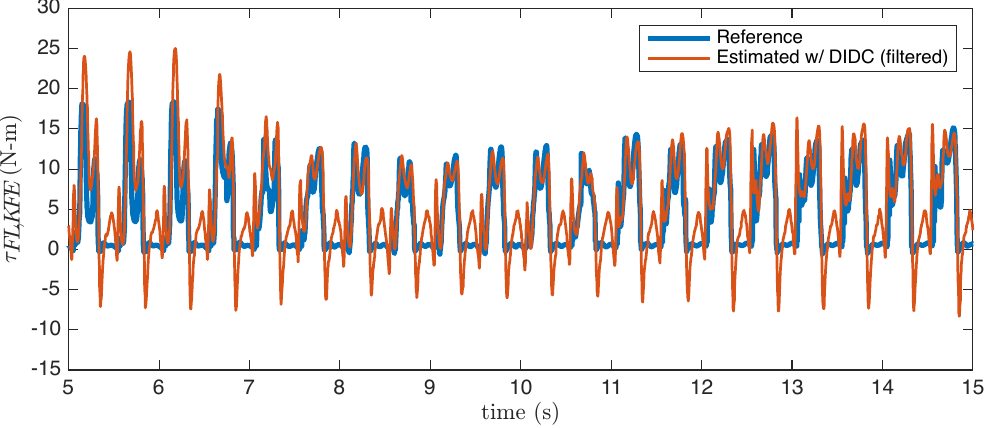}
    \caption{Estimated and reference torques of the FL leg joints on hardware.}
    \label{fig:torque_tracking}
\end{figure}

Since the NSPIDC controller does not work at high velocities, we chose the same movement routine for testing but with a reduced commanded velocity of $\pm0.5$ m/s in each direction. Fig. \ref{fig:velocity_tracking} shows the comparison of the reference velocity of the base against the estimated velocity from the state estimator.
The NSPIDC controller still has higher foot slip even at lower velocity, as shown in Fig. \ref{fig:hw_feet_vel}. The maximum foot velocity deviation for the experiment was $1.02$ m/s for NSPIDC as compared to $0.91$ m/s for DIDC. Primarily the foot slip is observed during the touchdown phase of the foot. The NSPIDC controller is unable to continue motion when the foot slip becomes considerably higher than the other controllers, as was shown in the simulation results. 
The reference and estimated torques for the first 10 seconds of motion using the DIDC controller are shown in Fig. \ref{fig:torque_tracking}. We show only 10 seconds of data (from 5 seconds to 15 seconds) for torques here for the sake of clarity.

\begin{figure}
    \centering
    \includegraphics[width=1.0\linewidth]{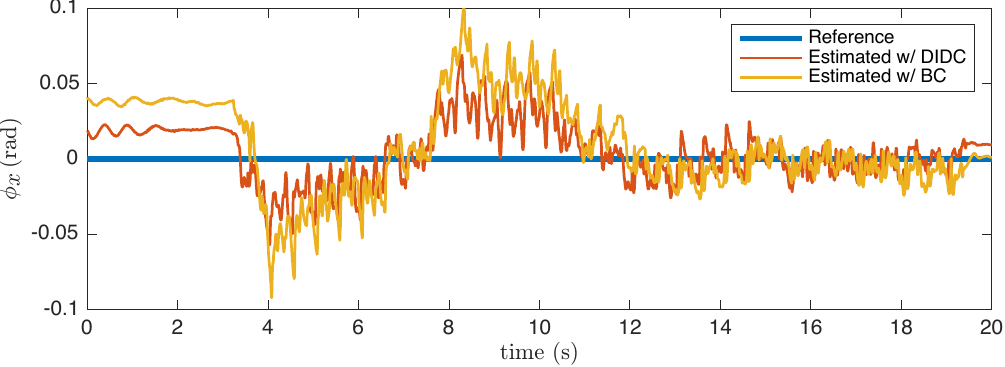}
    \includegraphics[width=1.0\linewidth]{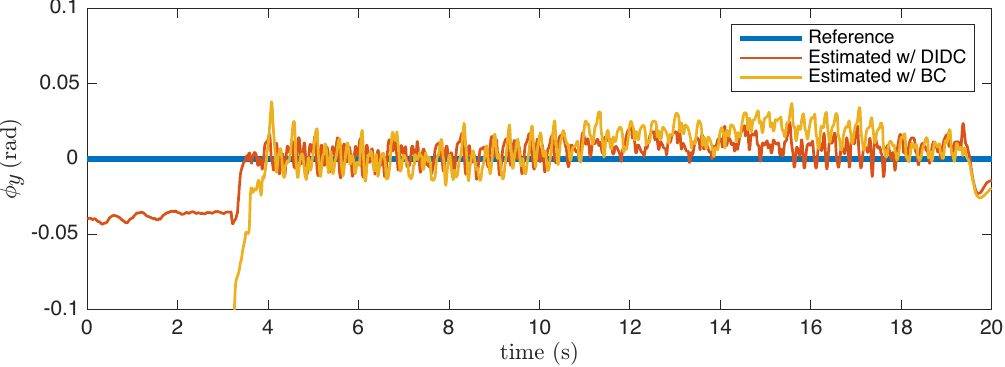}
    \caption{Estimated and reference roll and pitch angles of the robot base for the entire duration on hardware.}
    \label{fig:orientation_tracking}
\end{figure}

\begin{figure}
    \centering
    \includegraphics[width=1.0\linewidth]{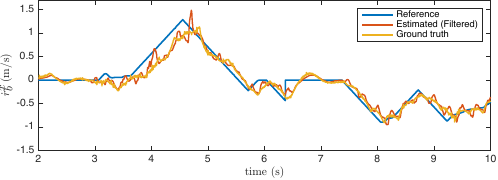}
    \includegraphics[width=1.0\linewidth]{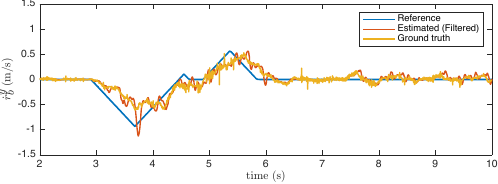}
    \caption{Comparison of estimated base velocity against ground truth data on hardware.}
    \label{fig:velocity_ground_truth}
\end{figure}

The estimated torques align closely with the reference torque generated by the controller as shown in Fig. \ref{fig:torque_tracking}. This shows that the generated torques are being tracked by the motor controller without significant contribution from the low-level PD controller and adds to the validity of the proposed feedback controller. During this motion, the variation in the roll and pitch angles of the robot base is shown in Fig. \ref{fig:orientation_tracking}.
The orientation errors during the entire motion remain within $0.1\ radians\ (5.72\degree)$ using both DIDC and BC. As discussed in Section II above, considering the effect of leg dynamics on base, and orthogonalizing base and joint tracking torques, improves base orientation tracking in DIDC as compared to balance controller. The average absolute orientation error in the graph above is $0.0234\ radians\ (1.34\degree)$ for BC while it is $0.0156\ radians\ (0.89\degree)$ for DIDC ($\sim 30$\% reduction in error). This also verifies the different power consumption behavior from Fig. \ref{fig:power_stats} since DIDC performs lesser work for reducing the errors, hence consumes lesser total power. 
To validate the state estimator, the estimated base velocity is compared against the ground truth data obtained from motion capture (mocap) system (Vicon Valkyrie). For this test, the robot is commanded using a joystick, and the position data from the mocap system is numerically differentiated to obtain the ground truth velocity of the robot base. The results are shown in Fig. \ref{fig:velocity_ground_truth}.
The state estimator is able to estimate the velocity of the robot with a mean absolute error of $\sim 0.066$m/s and a mean error of $\sim 0.00044$ m/s. This shows that the estimator has an accurate mean estimate and can be used for getting reliable velocity data for the controller.
\section{Conclusion}
\label{sec:conclude}
In this work, we have proposed a reactive controller that uses the full-RBD model of a quadruped robot and exact friction cone constraints to compute the required joint torque command, while being computationally lightweight. We achieve this by avoiding the dependence on general-purpose QP solvers and proposing a custom solver that exploits the geometry of the friction constraints. We compare the performance of our controller with the projection-based inverse dynamics controller and QP-based balance controllers in omnidirectional mobility routines on flat terrain, in simulations, and on hardware. The results reveal that the proposed controller improves the base motion tracking, reduces foot slippage, and consumes less power, especially at higher velocities. In future work, we plan to test our controller in the two-layer predictive-reactive control architecture. Also, further studies will investigate the nature of constraints, in addition to the cone constraint, that can be handled by the geometric projection-based approach used by our custom solver. 

\setcounter{secnumdepth}{0}

\begin{acks}
We thank the Mobile Robotics Laboratory (MRL) at IIT Kanpur and the Control/Robotics Research Laboratory (CRRL) at NYU for providing the resources to conduct the research and experiments for this paper.
\end{acks}

\section{Author Contributions}
N. Khandelwal led the development and implementation of the DIDC and GPGD algorithms, implemented baseline controllers, designed the software architecture, conducted experiments, and wrote the primary manuscript draft. A. Manu implemented the planning and estimation modules and the balance controller, contributed to GPGD debugging and optimization, and made substantial contributions to both the manuscript and control software implementation, including the safety state machine design. P. Krishnamurthy provided critical technical feedback on the manuscript and experimental methodology. F. Khorrami, M. Kothari, and S. S. Gupta provided editorial oversight and final manuscript revisions.

\bibliographystyle{SageH}
\bibliography{references.bib}

\begin{thebibliography}{34}
\providecommand{\natexlab}[1]{#1}
\providecommand{\url}[1]{\texttt{#1}}
\providecommand{\urlprefix}{URL }
\expandafter\ifx\csname urlstyle\endcsname\relax
  \providecommand{\doi}[1]{DOI:\discretionary{}{}{}#1}\else
  \providecommand{\doi}{DOI:\discretionary{}{}{}\begingroup \urlstyle{rm}\Url}\fi

\bibitem[{Aghili(2017)}]{qcqp-aghili}
Aghili F (2017) Quadratically constrained quadratic-programming based control of legged robots subject to nonlinear friction cone and switching constraints.
\newblock \emph{IEEE/ASME Transactions on Mechatronics} 22(6): 2469--2479.
\newblock \doi{10.1109/TMECH.2017.2755859}.

\bibitem[{Bledt et~al.(2018{\natexlab{a}})Bledt, Powell, Katz, Di~Carlo, Wensing and Kim}]{cheetah3}
Bledt G, Powell MJ, Katz B, Di~Carlo J, Wensing PM and Kim S (2018{\natexlab{a}}) Mit cheetah 3: Design and control of a robust, dynamic quadruped robot.
\newblock In: \emph{Proceedings of the 2018 IEEE/RSJ International Conference on Intelligent Robots and Systems (IROS)}. pp. 2245--2252.
\newblock \doi{10.1109/IROS.2018.8593885}.

\bibitem[{Bledt et~al.(2018{\natexlab{b}})Bledt, Wensing, Ingersoll and Kim}]{contact_estimation}
Bledt G, Wensing PM, Ingersoll S and Kim S (2018{\natexlab{b}}) Contact model fusion for event-based locomotion in unstructured terrains.
\newblock In: \emph{Proceedings of the 2018 IEEE International Conference on Robotics and Automation (ICRA)}. pp. 4399--4406.
\newblock \doi{10.1109/ICRA.2018.8460904}.

\bibitem[{Bloesch et~al.(2013)Bloesch, Hutter, Hoepflinger, Leutenegger, Gehring, Remy and Siegwart}]{bloesch}
Bloesch M, Hutter M, Hoepflinger MA, Leutenegger S, Gehring C, Remy CD and Siegwart R (2013) State estimation for legged robots: Consistent fusion of leg kinematics and imu.
\newblock In: \emph{Robotics: Science and Systems VIII}. The MIT Press.

\bibitem[{Buchli et~al.(2009)Buchli, Kalakrishnan, Mistry, Pastor and Schaal}]{idc_qr2}
Buchli J, Kalakrishnan M, Mistry M, Pastor P and Schaal S (2009) Compliant quadruped locomotion over rough terrain.
\newblock In: \emph{Proceedings of the 2009 IEEE/RSJ International Conference on Intelligent Robots and Systems}. pp. 814--820.
\newblock \doi{10.1109/IROS.2009.5354681}.

\bibitem[{Dario~Bellicoso et~al.(2016)Dario~Bellicoso, Gehring, Hwangbo, Fankhauser and Hutter}]{wbc}
Dario~Bellicoso C, Gehring C, Hwangbo J, Fankhauser P and Hutter M (2016) Perception-less terrain adaptation through whole body control and hierarchical optimization.
\newblock In: \emph{Proceedings of the 2016 IEEE-RAS 16th International Conference on Humanoid Robots (Humanoids)}. pp. 558--564.
\newblock \doi{10.1109/HUMANOIDS.2016.7803330}.

\bibitem[{Di~Carlo et~al.(2018)Di~Carlo, Wensing, Katz, Bledt and Kim}]{convex-mpc}
Di~Carlo J, Wensing PM, Katz B, Bledt G and Kim S (2018) Dynamic locomotion in the mit cheetah 3 through convex model-predictive control.
\newblock In: \emph{Proceedings of the 2018 IEEE/RSJ International Conference on Intelligent Robots and Systems (IROS)}. pp. 1--9.
\newblock \doi{10.1109/IROS.2018.8594448}.

\bibitem[{Doty et~al.(1993)Doty, Melchiorri and Bonivento}]{generalized_inverse}
Doty KL, Melchiorri C and Bonivento C (1993) A theory of generalized inverses applied to robotics.
\newblock \emph{The International Journal of Robotics Research} 12(1): 1--19.

\bibitem[{Fahmi et~al.(2019)Fahmi, Mastalli, Focchi and Semini}]{wbc-fahmi}
Fahmi S, Mastalli C, Focchi M and Semini C (2019) Passive whole-body control for quadruped robots: Experimental validation over challenging terrain.
\newblock \emph{IEEE Robotics and Automation Letters} 4(3): 2553--2560.
\newblock \doi{10.1109/LRA.2019.2908502}.

\bibitem[{Farshidian et~al.(2017{\natexlab{a}})Farshidian, Jelavić, Winkler and Buchli}]{imc}
Farshidian F, Jelavić E, Winkler AW and Buchli J (2017{\natexlab{a}}) Robust whole-body motion control of legged robots.
\newblock In: \emph{Proceedings of the 2017 IEEE/RSJ International Conference on Intelligent Robots and Systems (IROS)}. pp. 4589--4596.
\newblock \doi{10.1109/IROS.2017.8206328}.

\bibitem[{Farshidian et~al.(2017{\natexlab{b}})Farshidian, Neunert, Winkler, Rey and Buchli}]{nmpc-farbod}
Farshidian F, Neunert M, Winkler AW, Rey G and Buchli J (2017{\natexlab{b}}) An efficient optimal planning and control framework for quadrupedal locomotion.
\newblock In: \emph{Proceedings of the 2017 IEEE International Conference on Robotics and Automation (ICRA)}. pp. 93--100.
\newblock \doi{10.1109/ICRA.2017.7989016}.

\bibitem[{Ferreau et~al.(2014)Ferreau, Kirches, Potschka, Bock and Diehl}]{qpoases}
Ferreau HJ, Kirches C, Potschka A, Bock HG and Diehl M (2014) qpoases: A parametric active-set algorithm for quadratic programming.
\newblock \emph{Mathematical Programming Computation} 6: 327--363.

\bibitem[{Focchi et~al.(2017)Focchi, Del~Prete, Havoutis, Featherstone, Caldwell and Semini}]{vmc-slope}
Focchi M, Del~Prete A, Havoutis I, Featherstone R, Caldwell DG and Semini C (2017) High-slope terrain locomotion for torque-controlled quadruped robots.
\newblock \emph{Autonomous Robots} 41: 259--272.

\bibitem[{Foundation(2024)}]{cyclonedds}
Foundation E (2024) {Eclipse Cyclone DDS}.
\newblock \urlprefix\url{https://github.com/ eclipse-cyclonedds/cyclonedds}.
\newblock Accessed on [20-07-2024].

\bibitem[{Gehring et~al.(2013)Gehring, Coros, Hutter, Bloesch, Hoepflinger and Siegwart}]{vmc2013}
Gehring C, Coros S, Hutter M, Bloesch M, Hoepflinger MA and Siegwart R (2013) Control of dynamic gaits for a quadrupedal robot.
\newblock In: \emph{Proceedings of the 2013 IEEE International Conference on Robotics and Automation}. pp. 3287--3292.
\newblock \doi{10.1109/ICRA.2013.6631035}.

\bibitem[{Hafner et~al.(2019)Hafner, Lillicrap, Ba and Norouzi}]{learning1}
Hafner D, Lillicrap T, Ba J and Norouzi M (2019) Dream to control: Learning behaviors by latent imagination.
\newblock \emph{arXiv preprint arXiv:1912.01603} .

\bibitem[{Hildebrand(1989)}]{hildebrand}
Hildebrand M (1989) The quadrupedal gaits of vertebrates.
\newblock \emph{BioScience} 39(11): 766.

\bibitem[{Hutter et~al.(2016)Hutter, Gehring, Jud, Lauber, Bellicoso, Tsounis, Hwangbo, Bodie, Fankhauser, Bloesch, Diethelm, Bachmann, Melzer and Hoepflinger}]{anymal}
Hutter M, Gehring C, Jud D, Lauber A, Bellicoso CD, Tsounis V, Hwangbo J, Bodie K, Fankhauser P, Bloesch M, Diethelm R, Bachmann S, Melzer A and Hoepflinger M (2016) Anymal - a highly mobile and dynamic quadrupedal robot.
\newblock In: \emph{Proceedings of the 2016 IEEE/RSJ International Conference on Intelligent Robots and Systems (IROS)}. pp. 38--44.
\newblock \doi{10.1109/IROS.2016.7758092}.

\bibitem[{Jenelten et~al.(2024)Jenelten, He, Farshidian and Hutter}]{learning2}
Jenelten F, He J, Farshidian F and Hutter M (2024) Dtc: Deep tracking control.
\newblock \emph{Science Robotics} 9(86): eadh5401.

\bibitem[{Kalman(1960)}]{lqr}
Kalman RE (1960) Contributions to the theory of optimal control.
\newblock \emph{Bol. soc. mat. mexicana} 5(2): 102--119.

\bibitem[{Kang et~al.(2023)Kang, Cheng, Zamora, Zargarbashi and Coros}]{learning3}
Kang D, Cheng J, Zamora M, Zargarbashi F and Coros S (2023) Rl + model-based control: Using on-demand optimal control to learn versatile legged locomotion.
\newblock \emph{IEEE Robotics and Automation Letters} 8(10): 6619--6626.
\newblock \doi{10.1109/LRA.2023.3307008}.

\bibitem[{Kim et~al.(2019)Kim, Di~Carlo, Katz, Bledt and Kim}]{wbic-mit}
Kim D, Di~Carlo J, Katz B, Bledt G and Kim S (2019) Highly dynamic quadruped locomotion via whole-body impulse control and model predictive control.
\newblock \emph{arXiv preprint arXiv:1909.06586} .

\bibitem[{Liu et~al.(2016)Liu, Tan and Padois}]{hierarchical}
Liu M, Tan Y and Padois V (2016) Generalized hierarchical control.
\newblock \emph{Autonomous Robots} 40: 17--31.

\bibitem[{Mistry et~al.(2010)Mistry, Buchli and Schaal}]{idc_qr}
Mistry M, Buchli J and Schaal S (2010) Inverse dynamics control of floating base systems using orthogonal decomposition.
\newblock In: \emph{Proceedings of the 2010 IEEE International Conference on Robotics and Automation}. pp. 3406--3412.
\newblock \doi{10.1109/ROBOT.2010.5509646}.

\bibitem[{Neunert et~al.(2018)Neunert, Stäuble, Giftthaler, Bellicoso, Carius, Gehring, Hutter and Buchli}]{nmpc-neunert}
Neunert M, Stäuble M, Giftthaler M, Bellicoso CD, Carius J, Gehring C, Hutter M and Buchli J (2018) Whole-body nonlinear model predictive control through contacts for quadrupeds.
\newblock \emph{IEEE Robotics and Automation Letters} 3(3): 1458--1465.
\newblock \doi{10.1109/LRA.2018.2800124}.

\bibitem[{Raibert(1986)}]{raibert_book}
Raibert MH (1986) \emph{Legged robots that balance}.
\newblock MIT press.

\bibitem[{Raiola et~al.(2020)Raiola, Mingo~Hoffman, Focchi, Tsagarakis and Semini}]{wbc-simple}
Raiola G, Mingo~Hoffman E, Focchi M, Tsagarakis N and Semini C (2020) A simple yet effective whole-body locomotion framework for quadruped robots.
\newblock \emph{Frontiers in Robotics and AI} 7: 528473.

\bibitem[{Righetti et~al.(2011)Righetti, Buchli, Mistry and Schaal}]{idc_qr3}
Righetti L, Buchli J, Mistry M and Schaal S (2011) Inverse dynamics control of floating-base robots with external constraints: A unified view.
\newblock In: \emph{Proceedings of the 2011 IEEE International Conference on Robotics and Automation}. pp. 1085--1090.
\newblock \doi{10.1109/ICRA.2011.5980156}.

\bibitem[{Savitzky and Golay(1964)}]{savitzky_golay}
Savitzky A and Golay MJ (1964) Smoothing and differentiation of data by simplified least squares procedures.
\newblock \emph{Analytical chemistry} 36(8): 1627--1639.

\bibitem[{Siciliano et~al.(2009)Siciliano, Sciavicco, Villani and Oriolo}]{robot-book}
Siciliano B, Sciavicco L, Villani L and Oriolo G (2009) \emph{Robotics: Modelling, Planning, and Control}.
\newblock 1 edition. Springer London.

\bibitem[{Sleiman et~al.(2021)Sleiman, Farshidian, Minniti and Hutter}]{sleiman}
Sleiman JP, Farshidian F, Minniti MV and Hutter M (2021) A unified mpc framework for whole-body dynamic locomotion and manipulation.
\newblock \emph{IEEE Robotics and Automation Letters} 6(3): 4688--4695.
\newblock \doi{10.1109/LRA.2021.3068908}.

\bibitem[{Stellato et~al.(2020)Stellato, Banjac, Goulart, Bemporad and Boyd}]{osqp}
Stellato B, Banjac G, Goulart P, Bemporad A and Boyd S (2020) Osqp: An operator splitting solver for quadratic programs.
\newblock \emph{Mathematical Programming Computation} 12(4): 637--672.

\bibitem[{Tasora and Anitescu(2011)}]{friction_projection}
Tasora A and Anitescu M (2011) A matrix-free cone complementarity approach for solving large-scale, nonsmooth, rigid body dynamics.
\newblock \emph{Computer Methods in Applied Mechanics and Engineering} 200(5-8): 439--453.

\bibitem[{Todorov et~al.(2012)Todorov, Erez and Tassa}]{mujoco}
Todorov E, Erez T and Tassa Y (2012) Mujoco: A physics engine for model-based control.
\newblock In: \emph{Proceedings of the 2012 IEEE/RSJ International Conference on Intelligent Robots and Systems}. pp. 5026--5033.
\newblock \doi{10.1109/IROS.2012.6386109}.

\end{thebibliography}

\end{document}